\documentclass[conference]{IEEEtran}
\IEEEoverridecommandlockouts
\usepackage{cite}
\usepackage{amsmath,amssymb,amsfonts}
\usepackage{geometry}
\usepackage{graphicx}
\usepackage{textcomp}
\usepackage{xcolor}
\usepackage{subfigure}
\usepackage{algorithm}
\usepackage{algpseudocode}
\usepackage{caption}
\usepackage[justification=raggedright,singlelinecheck=false]{caption}
\usepackage{bbding}
\usepackage{times}
\usepackage{multirow}
\usepackage{caption}
\usepackage{booktabs}
\usepackage{amssymb}
\def\BibTeX{{\rm B\kern-.05em{\sc i\kern-.025em b}\kern-.08em
    T\kern-.1667em\lower.7ex\hbox{E}\kern-.125emX}}
\begin{document}
\title{Diffusion-Driven Semantic Communication for Generative Models with Bandwidth Constraints\\}

\author{\IEEEauthorblockN{Lei Guo, Wei Chen, \IEEEmembership{Senior Member, IEEE}, Yuxuan Sun, \IEEEmembership{Member, IEEE}, Bo Ai, \IEEEmembership{Fellow, IEEE}, \\
Nikolaos Pappas, \IEEEmembership{Senior Member, IEEE}, Tony Q. S. Quek, \IEEEmembership{Fellow, IEEE}}

\thanks{
Lei Guo, Wei Chen and Bo Ai are with the State Key Laboratory of Advanced Rail Autonomous Operation, Beijing Jiaotong University, China and the School of Electronic and Information Engineering, Beijing Jiaotong University, Beijing, China. Yuxuan Sun is with the School of Electronic and Information Engineering, Beijing Jiaotong University, Beijing, China. (e-mail:leiguo@bjtu.edu.cn; weich@bjtu.edu.cn; yxsun@bjtu.edu.cn; boai@bjtu.edu.cn). Corresponding author: Wei Chen, Yuxuan Sun.

Nikolaos Pappas is with the Department of Computer and Information Science, Linköping University, Linköping, Sweden (e-mail:nikolaos.pappas@liu.se).

Tony Quek is with the Information Systems Technology and Design, Singapore University of Technology and Design, Singapore, Singapore (e-mail:tonyquek@sutd.edu.sg).

This work was supported in part by the National Key R\&D Program of China under Grant 2024YFE0200700; the Natural Science Foundation of China (U2468201, W2421083, 62221001, 62301024); the Beijing Natural Science Foundation under grant L222044, and the Talent Fund of Beijing Jiaotong University under grant 2023XKRC030.}
}

% This work was supported in part by the Fundamental Research Funds for the Central Universities (2022JBQY004); the Natural Science Foundation of China (U2468201, 62122012, 62221001, 62301024); the Beijing Natural Science Foundation (L222044, L211012) and the Talent Fund of Beijing Jiaotong University under grant 2023XKRC030.
\maketitle

\begin{abstract}
%Generative artificial intelligence is widely used in various fields due to its remarkable capabilities. 
Diffusion models have been extensively utilized in AI-generated content (AIGC) in recent years, thanks to the superior generation capabilities. Combining with semantic communications, diffusion models are used for tasks such as denoising, data reconstruction, and content generation. However, existing diffusion-based generative models do not consider the stringent bandwidth limitation, which limits its application in wireless communication. This paper introduces a diffusion-driven semantic communication framework with advanced VAE-based compression for bandwidth-constrained generative model. Our designed architecture utilizes the diffusion model, where the signal transmission process through the wireless channel acts as the forward process in diffusion. To reduce bandwidth requirements, we incorporate a downsampling module and a paired upsampling module based on a variational auto-encoder with reparameterization at the receiver to ensure that the recovered features conform to the Gaussian distribution. Furthermore, we derive the loss function for our proposed system and evaluate its performance through comprehensive experiments. Our experimental results demonstrate significant improvements in pixel-level metrics such as peak signal to noise ratio (PSNR) and semantic metrics like learned perceptual image patch similarity (LPIPS). These enhancements are more profound regarding the compression rates and SNR compared to deep joint source-channel coding (DJSCC). We release the code at https://github.com/import-sudo/Diffusion-Driven-Semantic-Communication.
\end{abstract}
%Experimental results demonstrate that our digital system can achieve less than 1\% performance degradation in 6dB signal-to-noise ratios (SNRs). 

\section{Introduction}

Deep learning-based semantic communication has recently demonstrated significant advantages for next-generation wireless communications, particularly in task-oriented communication systems designed to perform specific tasks or achieve particular objectives by extracting and transmitting only the relevant semantic information necessary. This approach has been applied to images \cite{9998051,9653664,10772628}, texts \cite{2021Deep}, and speech signals \cite{2021Semantic}. Neural networks can effectively extract useful information and handle complex data by learning the underlying features of the data. Numerous works have focused on improving communication efficiency. For example, deep joint source-channel coding (DJSCC)\cite{9438648} merges source coding and channel coding into a single optimization problem with the goal of minimizing the overall transmission rate required to achieve a specific level of reliability. Compared to traditional wireless communication systems, semantic communication focuses on transmitting information in the semantic domain \cite{deng2023deep}. Its primary goal is to ensure that the intended information of the message is accurately conveyed to the receiver. 

Generative AI \cite{10273408,10172151} is becoming an essential application at the wireless edge. The generation models in generative AI have applications across various domains, including image, text, and audio generation. In image generation, these models can create new images based on given prompts or conditional inputs, effectively mapping latent features extracted by the model, to images. In wireless networks, image generation plays a crucial role in applications such as augmented reality (AR), virtual reality (VR), and mobile gaming through high-quality real-time visuals. Efficient image generation at the resource-limited devices reduces bandwidth usage and latency, vital for advanced applications like remote education and smart city surveillance. In semantic communication, the model leverages the advantages of semantic extraction and representation to transmit information at the semantic level \cite{gao2024intelligence}. This process can also be seen as a conditional generation task under given channels and image features, i.e., generating corresponding images based on given conditions. Generative models over wireless networks benefit from semantic communication by focusing on transmitting essential features or information instead of raw data, which reduces data volume and enhances bandwidth efficiency. This approach also decreases latency, making it suitable for real-time applications, and improves robustness against noise and signal degradation.

%\cite{10001359, 10158995, 10233814, choi2019neural, 9791398, 10437849, 10480348, letafati2023probabilistic, 10304477, 10448462, yilmaz2023high}
Several works aim to leverage the semantic extraction capabilities of the generative models for semantic communication. Human visual perception in semantic communication is explored in \cite{10001359}, where generative adversarial networks (GANs) capture global semantic information and local textures, producing images resembling human visual perception. Additionally, GenerativeJSCC, introduced in \cite{10158995}, utilizes Style-GAN \cite{9156570} to enhance quality in edge cases. The lightweight GenerativeJSCC is proposed in \cite{10233814}, reducing the computation of the generation network while maintaining performance. A novel discrete variational auto-encoder (VAE) model is proposed in \cite{choi2019neural} for semantic communication, while an adaptive rate transmission based on VAE is further explored in \cite{9791398}. 

Recently, more work has focused on diffusion models for reducing the impact of wireless noise. The diffusion model \cite{ho2020denoising,9878449} adds Gaussian white noise during the forward process and gradually denoises the image during the backward process. This process can fit the communication system in an Additive White Gaussian Noise (AWGN) channel. For example, in \cite{10437849} and \cite{10480348}, the diffusion model is employed to remove channel noise and improve reconstruction quality. The denoising diffusion probabilistic model (DDPM) is further applied in \cite{letafati2023probabilistic} for probabilistic constellation shaping in wireless communications. To address perception distortion in finite block lengths, diffusion is integrated into DJSCC in \cite{yilmaz2023high}. Moreover, some works use degraded signals based on diffusion models for semantic communication. A hybrid joint source-channel coding (JSCC) scheme is proposed in \cite{10304477}, where conventional digital communication with the compressed image is complemented with a generative refinement component to enhance the perceptual quality of reconstruction. Invertible Neural Networks are proposed in \cite{10448462}, where the signal is decomposed at the transmitter and is estimated from the degraded part using the diffusion models to recover high-quality source images under extreme conditions at the receiver. The diffusion models are used to denoise in wireless communication. However, current diffusion-based approaches have not fully harnessed the generative capabilities of pre-trained diffusion models \cite{9878449}, which were extensively trained on the LAION-5B dataset \cite{NEURIPS2022_a1859deb} with billions of images that required immense GPU resources. Although integrating these pre-trained diffusion models helps preserve their powerful generative capabilities, it presents challenges in flexibly adapting to varying bandwidth constraints and limits their adaptability.

% For example, the problems in federated learning (FL) focus on the limited communication resource and convergence \cite{10024766, pmlr}. Due to limited communication bandwidth, the digital scheme is first proposed in federated learning over wireless fading channels, where the sparsified gradient is utilized for bandwidth reduction \cite{9014530}. To improve communication efficiency, an online device scheduling policy for over-the-air FL is proposed in \cite{9605599}. Additionally, a novel distributed compression framework is proposed with independent encoders and a joint decoder, which can flexibly compress data from multiple sources to adapt any available bandwidth \cite{Li2023taskaware}. To improve the convergence speed, device selection and beamforming are designed for over-the-air FL in \cite{8952884}. In \cite{10145043}, a new metric is proposed to minimize the difference between the aggregated gradient of scheduled devices and the full participation gradient. 
There are several works exploring bandwidth-compression semantic communication. Some of them focus on VAE and diffusion models for compression. The latent variable model with a quantization-aware posterior and prior is designed in \cite{10030851,10274142} for lossy image compression, where a hierarchical VAE architecture is integrated. Several works focus on specific tasks using VAE compression, combining the VAE decoder and task-oriented detector into a compressed task detector. A bridge network is proposed in \cite{9675387} to adopt for learning a compact representation. A VAE-based joint compression and classification model is proposed in \cite{9794700}, which enables learning on the latent feature space to efficiently encode/compress and effectively classify images through end-to-end training. There are also several research works that explore compression methods based on diffusion models. An additional latent variable is utilized as a condition on the reverse diffusion process in \cite{2024Lossy}. A novel diffusion-driven image compression framework with a privileged end-to-end decoder is proposed in \cite{2024Correcting}, where the privileged decoder helps correct the sampling process with only a few bits to achieve better reconstruction. Stable diffusion is proposed in \cite{9878449}, where the latent space is applied with cross-attention layers to facilitate diffusion model training on limited computational resources. These compression works based on VAE and diffusion focus on reducing computational resources. In addition, there are other works leveraging unstructured data with graph neural network (GNN) \cite{2016Semi,9046288} to explore bandwidth compression. For example, in \cite{9814464}, the modulated symbols of users are treated as nodes and applied to a graph neural network to mitigate multi-user interference and reduce the bandwidth required for transmission while achieving the desired classification accuracy. Additionally, a pragmatic semantic communication framework based on GNNs is proposed in \cite{10454639}, where a semantic feedback level is introduced to provide information on the perceived semantic effectiveness with minimal overhead. 
%However, in semantic communication scenarios, both computation and bandwidth are limited and constrained, especially in edge computing environments.
%In \cite{zheng2024genet}, GNN-based semantic communication is utilized for SNR-unaware scenarios, enhancing the robustness and adaptability of the communication system.

\begin{figure*}
\centering
\includegraphics[width=\linewidth]{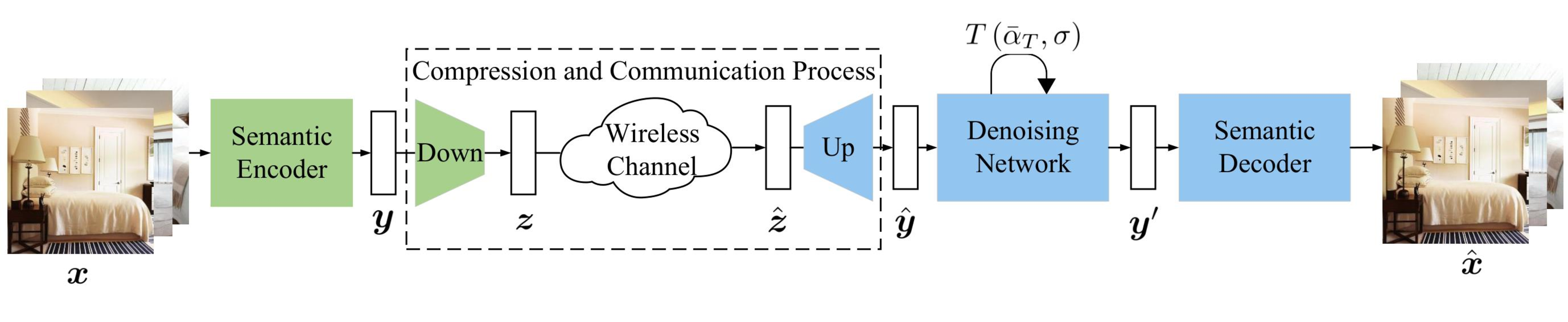}
\caption{The framework of the diffusion-driven communication-efficient semantic communication system.} %The encoder, diffusion and decoder are utilized from those of the stable diffusion model \cite{9878449}.}
\label{framework_diffusion}
\end{figure*}

%%%%%%%%
%%%%%%%%
% VAE-based comression/ why compression based on this framework
In this paper, we propose a method where the received signal is transformed into the Gaussian-distributed features through forward process in diffusion models, with varying SNRs mapped to different diffusion steps $T$. The received signal is then processed through $T$ iterations of reverse denoising through the Stable Diffusion model to mitigate channel noise. To reduce transmission bandwidth while leveraging the remarkable generative capabilities of pre-trained diffusion models, we integrate a downsampling module for compression and a VAE-based upsampling module at the receiver for reconstruction. The VAE-based module ensures the Gaussian conditions required by the diffusion model and incorporates SNR as a conditioning input to address the sensitivity of feature variance to channel noise. Our contributions can be summarized as follows:

\begin{itemize}
\item[$\bullet$]We propose a communication-efficient generative semantic communication system, which incorporates the forward process of diffusion and channel noise, mapping the channel noise into the $T$-th forward processes in diffusion, adaptable to different signal to noise ratios (SNRs) of the wireless channel. At the receiver, we effectively utilize the diffusion model of the reverse process to remove the channel noise. Multiple kinds of wireless channels, including AWGN, Rayleigh, and multiple-input multiple-output (MIMO) channels, are considered. For the Rayleigh channel, we combine our approach with mean squared error (MSE) equalization, while singular value decomposition (SVD) is utilized for the MIMO channel.

\item[$\bullet$] To reduce bandwidth requirements, we integrate a downsampling module to compress semantic features and a VAE-based upsampling network with reparameterization at the receiver, serving as a plug-and-play module in diffusion-based networks. This approach ensures that the reconstructed features preserve the Gaussian distribution. The VAE-based upsampling module is further designed by incorporating the SNR as a conditioning input for more accurate variance estimation.

\item[$\bullet$]To further enhance the feature extraction capability of the downsampling module and the recovery capability of the upsampling module, we integrate a guidance approach within this architecture that learns from the distribution of generators without bandwidth compression. Additionally, we introduce a comprehensive loss function that combines VAE-based compression and guidance. This loss function integrates contributions from both the VAE loss and the guidance loss, utilizing Kullback-Leibler (KL) divergence to effectively align the distributions of networks employing paired downsampling and upsampling modules with those lacking such modules.

\item[$\bullet$]Our experimental results demonstrate the effectiveness of the proposed architecture under bandwidth constraints. Our approach exhibits substantial improvements across various compression rates and SNRs compared to the baseline of the DJSCC-based downsampling and upsampling module. Specifically, the reparameterization significantly enhances pixel-level performance, while guidance is crucial in improving semantic transmission. %This comprehensive approach highlights the potential for advancing semantic communication systems, providing enhanced reliability and efficiency in generation performance.
\end{itemize}

The rest of this paper is organized as follows. Section II presents the system model for the proposed semantic communication system. In Section III, we propose the semantic diffusion-driven communication system, which integrates the forward process of diffusion and channel noise and is adaptable to various SNRs of the wireless channel. Section IV details the VAE-based semantic compression module and the comprehensive loss function. The effectiveness of the proposed methods is evaluated in Section V to demonstrate their validity. Finally, Section VI concludes our work.

\emph{Notations}: Vectors are denoted by boldface lower-case letters, such as $\boldsymbol{x}$. The symbol $\cdot$ denotes multiplication between real numbers and vectors, while the symbol $\otimes$ represents the multiplication of vectors by multiplying their elements at corresponding positions. $\mathbb{C}$ denotes the sets of complex numbers. $\mathcal{N}\left(\mu, \sigma \right)$ represents a Gaussian distribution with mean $\mu$ and variance $\sigma$.

\begin{figure*}
\centering
\includegraphics[width=0.98\linewidth]{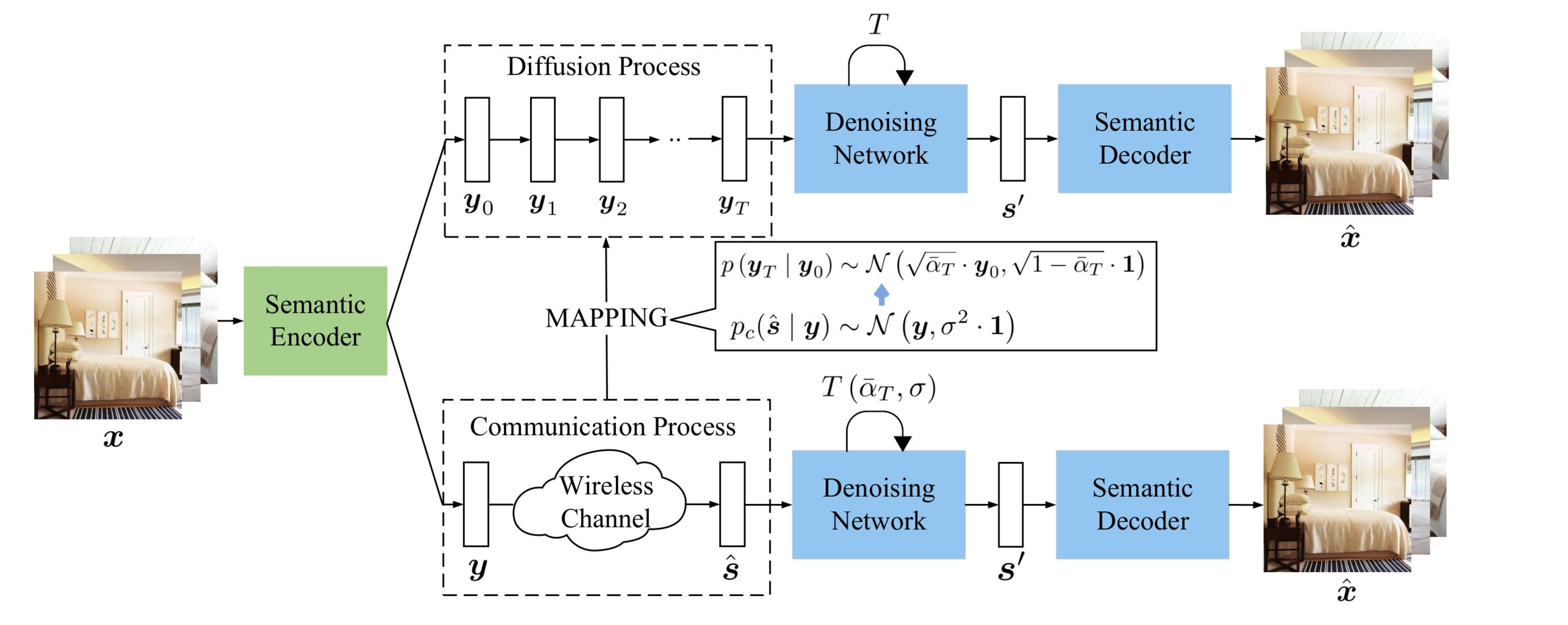}
\caption{The framework of the semantic encoder and decoder based on diffusion.} %The encoder, diffusion and decoder are utilized from those of the stable diffusion model \cite{9878449}.}
\label{teacher}
\end{figure*}

\section{System Model of diffusion-driven Communication-Efficient Semantic Communication System}
% diffusion-driven bandwidth-efficient semantic communication system
% diffusion-driven Semantic Communication for Bandwidth-Constrained Generative Tasks
% communication-efficient generative semantic communication system
In this section, we introduce the proposed communication-efficient generative semantic communication system based on diffusion and bandwidth compression. This framework benefits from the semantic extraction capability of Stable Diffusion \cite{9878449} and the efficiency of bandwidth compression to enhance the overall communication process. The proposed generative semantic communication system based on a diffusion model, is depicted in Fig. \ref{framework_diffusion}. Our focus is on image generation over an AWGN channel. Let $\boldsymbol{x}\in \mathbb{C}^{3WH}$ represents the vector reshaped from the input image, where $W$ and $H$ denote the width and height of the image, respectively. The semantic encoder transforms the input into latent features, which can be represented as:
\begin{equation}
\boldsymbol{y} = E(\boldsymbol{x};\boldsymbol{\theta}),
\end{equation}
%\in \mathbb{C}^{k}
where $\boldsymbol{y} \in \mathbb{C}^{whc}$ is the vector reshaped from semantic feature. $E(\cdot;\boldsymbol{\theta})$ denotes the semantic encoder with the parameter $\boldsymbol{\theta}$. To reduce the bandwidth for transmitting semantic features, we design a low-complexity downsampling network, which can be represented as:
\begin{equation}
\boldsymbol{z} = F_d(\boldsymbol{y};\boldsymbol{\psi}), 
\end{equation}
where $\boldsymbol{z} \in \mathbb{C}^{kwhc}$ is the transmitted complex-valued vector reshaped from semantic feature, with $k\in (0,1)$ being the compression factor. $F_d(\cdot;\boldsymbol{\psi})$ denotes the downsampling network with the parameter $\boldsymbol{\psi}$. The compression feature is transmitted over the channel, which is expressed as:
\begin{equation}
\hat{\boldsymbol{z}} = \boldsymbol{z} + \boldsymbol{n}_0,
\label{channel}
\end{equation}
where $\hat{\boldsymbol{z}}$ is the vector reshaped from received feature with channel noise following Gaussian distribution with variance $\sigma^{2}$, i.e., $\boldsymbol{n}_0 \sim  \mathcal {N}\left(\boldsymbol{0}, \sigma^2  \cdot \boldsymbol{1}\right)$. $\boldsymbol{0}$ and $\boldsymbol{1}$ are vectors where each element equals zero and one, respectively. The transmission process can be represented as a conditional probability distribution, which follows a Gaussian distribution, i.e., $p_c\left(\hat{\boldsymbol{z}} \mid \boldsymbol{z}\right) \sim \mathcal{N}\left(\boldsymbol{z}, \sigma^2 \cdot \boldsymbol{1}\right)$. 

At the receiver, the compression feature is recovered by the paired upsampling network. Additionally, to ensure the recovered features follow a Gaussian distribution, we utilize the VAE-based upsampling network to reconstruct the features, which can be represented as:
\begin{equation}
(\boldsymbol{\mu}_{y}, \boldsymbol{\sigma}_{y}) = F_u(\hat{\boldsymbol{z}}, \text{SNR};\boldsymbol{\omega}), 
\end{equation}
\begin{equation}
\hat{\boldsymbol{y}} = \boldsymbol{\mu}_{y}+ \boldsymbol{\sigma}_{y} \otimes \boldsymbol{\epsilon}_{y},
\end{equation}
where $\boldsymbol{\mu}_{y}, \boldsymbol{\sigma}_{y}, \boldsymbol{\epsilon}_{y} \in \mathbb{C}^{whc}$, $\boldsymbol{\epsilon}_y \sim  \mathcal {N}\left(\boldsymbol{0}, \boldsymbol{1}\right)$. $F_{u}(\cdot; \boldsymbol{\omega})$ denotes the upsampling network with the parameter $\boldsymbol{\omega}$. Since the reconstructed semantic feature is influenced by the noise of the wireless channel, we also incorporate the SNR as an input. We assume $\sigma$ is known at the the receiver, and $\text{SNR}=\frac{1}{kwhc\sigma^2}\left\|\boldsymbol{z}\right\|_2^2$. $\hat{\boldsymbol{y}} \sim  \mathcal {N}\left(\boldsymbol{\mu}_{y}, \boldsymbol{\sigma}_{y}^2\right)$ is the vector reshaped estimated semantic feature, which is denoised by the reverse process of diffusion with the diffusion parameter $\boldsymbol{\phi}$, i.e., $p_d\left(\boldsymbol{y}^{\prime} \mid \hat{\boldsymbol{y}}, \sigma^2, \boldsymbol{\phi} \right)$. The decoder transforms the reshaped vector $\boldsymbol{y}^{\prime}$ from diffusion into the image, which can be represented as:
\begin{equation}
\hat{\boldsymbol{x}} = D(\boldsymbol{y}^{\prime};\boldsymbol{\delta}),
\end{equation}
where $D(\cdot;\boldsymbol{\delta})$ denotes the semantic decoder with the parameter $\boldsymbol{\delta}$. In our framework, the parameters of encoder $\boldsymbol{\theta}$, diffusion $\boldsymbol{\phi}$ and decoder $\boldsymbol{\delta}$ are utilized in the same as those of the Stable Diffusion \cite{9878449}.

To reduce the distribution errors caused by the bandwidth compression module, we introduce an additional guidance-based loss during training, in addition to the VAE-based loss function. The overall loss function for the proposed diffusion-driven model with bandwidth compression and guidance can be expressed as follows:
\begin{equation}
\mathcal{L}(\boldsymbol{\psi}, \boldsymbol{\omega}, \boldsymbol{\phi}) = \mathcal{L}_{v} + \gamma \mathcal{L}_{g},
\end{equation}
where $\mathcal{L}_{v}$ and $\mathcal{L}_{g}$ represent the VAE-based loss function and the guidance-based loss function, respectively. $\gamma$ is a hyperparameter that controls the balance between the two components. This hybrid loss function will be introduced in detail in Section IV-C.
 
%The proposed compression and reconstruction modules not only effectively reduce the bandwidth requirements for data transmission but also ensure that the features follow a Gaussian distribution, which is required by the diffusion process. Additionally, the guidance from the teacher generator provides valuable information to assist the upsampling network of the student generator in better reconstructing the semantic features. This guidance also helps us train the downsampling and upsampling networks more efficiently, reducing computational training costs while maintaining good communication quality.

\section{Design of Adaptive Forward Process based on Diffusion}
In this section, we introduce the semantic encoder and decoder based on the adaptive forward process of diffusion, as shown in Fig. \ref{teacher}. In this design, we map the signal transmission process through the channel to the forward process of diffusion and utilize its reverse process to eliminate the noise. Additionally, this design serves as a crucial component in training the proposed framework with compression generator by providing valuable information and guidance.

\subsection{Channel as Part of the Forward Process}

For diffusion, the forward process involves gradually adding increasing levels of Gaussian noise to the data with a decreasing variance
schedule $\alpha_1, \alpha_2, \ldots, \alpha_T$, which can be expressed as:
\begin{equation}
p\left(\boldsymbol{y}_T \mid \boldsymbol{y}_{T-1}\right) \sim \mathcal{N}\left(\sqrt{\alpha_T} \cdot \boldsymbol{y}_{T-1}, \sqrt{1-\alpha_T} \cdot \boldsymbol{1}\right), 
\end{equation}
\begin{equation}
p\left(\boldsymbol{y}_T \mid \boldsymbol{y}_{0}\right) \sim \mathcal{N}\left(\sqrt{\bar{\alpha}_T} \cdot \boldsymbol{y}_{0}, \sqrt{1- \bar{\alpha}_T} \cdot \boldsymbol{1}\right),
\end{equation}
%\begin{equation}
%p_c(\hat{\boldsymbol{s}} \mid \boldsymbol{y}) \sim \mathcal{N}\left(\boldsymbol{y}, \sigma^2 \cdot \mathbf{1}\right), \boldsymbol{y}_0, \boldsymbol{y}_1, \boldsymbol{y}_2, T\left(\bar{\alpha}_T, \sigma\right)
%\end{equation}
where $\bar{\alpha}_T=\prod_{s=1}^T\alpha_s$, $\boldsymbol{y}_{0}=\boldsymbol{y}$. $\boldsymbol{y}_T$ can be sampled directly from the conditional probability, that is: 
\begin{equation}
\boldsymbol{y}_T = \sqrt{\bar{\alpha}_T} \cdot \boldsymbol{y}_0 + \sqrt{1- \bar{\alpha}_T} \cdot \boldsymbol{\epsilon}_{T},
\end{equation} 
where $\boldsymbol{\epsilon}_{T} \sim \mathcal{N}(\boldsymbol{0}, \boldsymbol{1})$. Note that $\bar{\alpha}_T$ is a hyperparameter that has been given in the diffusion-driven generation model.

The reverse process of diffusion involves reconstructing the distribution of $\boldsymbol{y}_{T-1}$ from $\boldsymbol{y}_T$. The deep learning network with parameters $\boldsymbol{\phi}$, such as U-Net, is employed to predict the posterior distribution, which can be represented as: 
\begin{equation}
 p_{\boldsymbol{\phi}}\left(\boldsymbol{y}_{T-1} \mid \boldsymbol{y}_T\right) \sim \mathcal{N}\left( \boldsymbol{\mu}_{\boldsymbol{\phi}}\left(\boldsymbol{y}_T, T\right), \boldsymbol{\Sigma}_{\boldsymbol{\phi}}\left(\boldsymbol{y}_T, T\right)\right),
\label{reverse}
\end{equation} 
\begin{equation}
 \boldsymbol{\mu}_{\boldsymbol{\phi}}\left(\boldsymbol{y}_T, T\right) =\frac{1}{\sqrt{\alpha_T}}\left(\boldsymbol{y}_T-\frac{1-\alpha_T}{\sqrt{1-\bar{\alpha}_T}} \cdot \boldsymbol{z}_{\boldsymbol{\phi}}\left(\boldsymbol{y}_T, T\right)\right),
\end{equation} 
\begin{equation}
 \boldsymbol{\Sigma}_{\boldsymbol{\phi}}\left(\boldsymbol{y}_T, T\right) =\frac{\left(1-\bar{\alpha}_{T-1}\right)\left(1-\alpha_T\right)}{1-\bar{\alpha}_T} \cdot \boldsymbol{1},
\end{equation} 
where $\boldsymbol{z}_{\boldsymbol{\phi}}\left(\boldsymbol{y}_T, T\right)$ is the predicted distribution of noise by the network given $\boldsymbol{y}_T$ and $T$.

As shown in Fig. \ref{teacher}, when the signal $\boldsymbol{y}$ is transmitted over the channel with the Gaussian noise of variance $\sigma^2$, it also can be represented as $\hat{\boldsymbol{s}} = \boldsymbol{y} + \sigma \cdot \boldsymbol{\epsilon}$, $\boldsymbol{\epsilon} \sim \mathcal{N}(\boldsymbol{0}, \boldsymbol{1})$. The received signal follows a Gaussian distribution with mean $\boldsymbol{y}$ and variance $\sigma^2 \cdot \boldsymbol{1}$. The transmission process through the wireless channel can be transformed into a forward process of diffusion. At the receiver, the reverse process of diffusion can be used to eliminate the noise and recover the signal. 

During the forward process of diffusion, a fixed step of Gaussian noise is added, typically around 200 steps in stable diffusion in the LSUN dataset, corresponding to an SNR of approximately -1dB. To simulate the noise in the forward process, the received feature with channel noise might be compensated by the remaining noise required in the forward process from the receiver, when the SNR of the channel is greater than -1dB. Under such circumstances, the impact of channel noise on the signal can be reduced as much as possible. To align the noise of the forward process and estimate the compensated noise at the receiver, $\boldsymbol{y}_T$ can be rewritten as:
\begin{equation}
\begin{aligned}
\boldsymbol{y}_T = & \sqrt{\bar{\alpha}_T} \cdot \left(\boldsymbol{y}_0 + \sqrt{\frac{1- \bar{\alpha}_T}{\bar{\alpha}_T}} \cdot \boldsymbol{\epsilon}_{T}\right) \\
%= & \sqrt{\bar{\alpha}_T} \cdot \left(\boldsymbol{y}_0 + \underbrace{\sigma \cdot \boldsymbol{\epsilon} + \sqrt{\frac{1- \bar{\alpha}_T}{\bar{\alpha}_T} - \sigma^2} \cdot \boldsymbol{\epsilon}^{\prime}}_{ \sqrt{\frac{1- \bar{\alpha}_T}{\bar{\alpha}_T}} \cdot \boldsymbol{\epsilon}_{T}}\right) \\
= & \sqrt{\bar{\alpha}_T} \cdot \left(\underbrace{\boldsymbol{y}_0 + \sigma \cdot \boldsymbol{\epsilon}}_{\hat{\boldsymbol{s}}} + \underbrace{\sqrt{\frac{1- \bar{\alpha}_T}{\bar{\alpha}_T} - \sigma^2} \cdot \boldsymbol{\epsilon}^{\prime}}_{\boldsymbol{n}_{\text{cps}}}\right),
\end{aligned}
\end{equation}
where $\boldsymbol{\epsilon}$ and $\boldsymbol{\epsilon}^{\prime}$ are independent Gaussian distributions, $\boldsymbol{\epsilon}^{\prime} \sim \mathcal{N}(\boldsymbol{0}, \boldsymbol{1})$. $\boldsymbol{n}_{\text{cps}}$ denotes the compensated noise. Thus, when $\bar{\alpha}_T<1 /\left(1+\sigma^2\right)$, the compensated noise is added to match the forward process to simulate the noise in the forward process at the receiver.
\subsection{Adaptive Forward Process}
The compensated noise at the receiver impacts the retrieval of detailed information from the received signal. To accommodate diverse channel conditions, such as when the channel quality is poor, the semantic content of transmitted information can be preserved. Conversely, more detailed information can be transmitted when the channel quality is better. Thus, we propose an adaptive forward process based on diffusion and wireless channel noise. When the channel quality is better, i.e., $\bar{\alpha}_T<1 /\left(1+\sigma^2\right)$, the compensated noise is removed at the receiver to allow for the reception of more detailed information. By treating the noise from the wireless channel as the full noise in a complete forward process, the loss of details is prevented. To transform the communication process into the diffusion process, the noise from the wireless channel is treated as the noise of the forward process, i.e., $p_c\left(\hat{\boldsymbol{s}} \mid \boldsymbol{y}\right) \sim p\left(\boldsymbol{y}_{u} \mid \boldsymbol{y}_0, \bar{\alpha}_{u} \right)$. This can be expressed as:
\begin{equation}
\begin{aligned}
\hat{\boldsymbol{s}} & = \boldsymbol{y} + \sigma \cdot \boldsymbol{\epsilon} \\
& =  \frac{1}{\sqrt{\bar{\alpha}_{u}}} \cdot \left(\sqrt{\bar{\alpha}_{u}} \cdot \boldsymbol{y}_0 + \frac{\sqrt{\bar{\alpha}_{u}}\sigma}{\sqrt{1-\bar{\alpha}_{u}} } \sqrt{1-\bar{\alpha}_{u}} \cdot \boldsymbol{\epsilon}\right).
\end{aligned}
\label{transfor}
\end{equation}

According to Eqn. (\ref{transfor}), when $|\frac{\sqrt{\bar{\alpha}_u} \sigma}{\sqrt{1-\bar{\alpha}_u}}| = 1$, i.e., $\bar{\alpha}_{u}=\frac{1}{\left(1+\sigma^2\right)}$, the transmission process over the channel is equivalent to the $u$-th step forward process. This can be expressed as:
\begin{equation}
\left. p_c\left(\hat{\boldsymbol{s}} \mid \boldsymbol{y} \right) \sim p\left(\boldsymbol{y}_{u} \mid \boldsymbol{y}_0 \right) \right|_{\bar{\alpha}_{u}=\frac{1}{\left(1+\sigma^2\right)}}.
\end{equation}
%\frac{\sqrt{\bar{\alpha}_{u}}\sigma}{\sqrt{1-\bar{\alpha}_{u}} } \sqrt{1-\bar{\alpha}_{u}}

The received signal $\hat{\boldsymbol{s}}$ is denoised by the reverse process of diffusion, expressed as:
\begin{equation}
\left. p_d\left(\hat{\boldsymbol{s}}_{u-1} \mid \hat{\boldsymbol{s}}_u, \boldsymbol{\phi} \right) \sim p_{\boldsymbol{\phi}}\left(\boldsymbol{y}_{u-1} \mid \boldsymbol{y}_{u}\right) \right|_{\bar{\alpha}_{u}=\frac{1}{\left(1+\sigma^2\right)}},
\end{equation}  
where $\hat{\boldsymbol{s}}_{u}$ denotes the $u$-th step feature of $\hat{\boldsymbol{s}}$ in the reverse process of diffusion. Here, $\hat{\boldsymbol{s}}_{u}=\hat{\boldsymbol{s}}$, $\hat{\boldsymbol{s}}_{0}=\boldsymbol{s}^{\prime}$. By transforming the noise of the wireless channel into the noise of the forward process, the impact of channel noise can be minimized. %Additionally, the distribution of $\hat{\boldsymbol{s}}$ within the framwork serves as guidance information to direct the training of the compression diffusion-driven semantic communication system, which is detailed in the next section.

\subsection{Transformation over Rayleigh and MIMO Channels}
To further evaluate the robustness and adaptability of our proposed framework, we extend the transformation of received signals into appropriate features for the forward process of diffusion models to more complex wireless channels, including Rayleigh and MIMO channels.

For the Rayleigh fading channel, MSE equalization is used to transform the received signals into Gaussian-distributed features suitable for the diffusion process. The received signal before equalization is given as:
\begin{equation}
\hat{\boldsymbol{s}} = \boldsymbol{h} \boldsymbol{y} + \sigma \cdot \boldsymbol{\epsilon},
\end{equation}
where $\boldsymbol{h}$ is the channel gain matrix. Using MSE equalization \cite{10480348}, the equalized received signal $\boldsymbol{s}_{\text{MSE}}$ is expressed as:
\begin{equation}
\boldsymbol{s}_{\text{MSE}} = \boldsymbol{w}_{\text{MSE}} \hat{\boldsymbol{s}}.
\end{equation}
The MSE equalization coefficient $\boldsymbol{w}_{\text{MSE}}$ is defined as:
\begin{equation}
\boldsymbol{w}_{\text{MSE}}=\frac{\boldsymbol{h}^*}{|\boldsymbol{h}|^2+\frac{\sigma^2}{|\boldsymbol{y}|^2}},
\end{equation}
where $\boldsymbol{h}^*$ is the conjugate of $\boldsymbol{h}$. Thus, the received signal after equalization is expressed as:
\begin{equation}
\begin{aligned}
\boldsymbol{s}_{\text{MSE}} & =\frac{|\boldsymbol{h}|^2}{|\boldsymbol{h}|^2+\frac{\sigma^2}{|\boldsymbol{y}|^2}} \cdot \boldsymbol{y}+ \sigma \cdot \frac{\boldsymbol{h}^*}{|\boldsymbol{h}|^2+\frac{\sigma^2}{|\boldsymbol{y}|^2}} \boldsymbol{\epsilon} \\
& =\frac{|\boldsymbol{h}|^2}{|\boldsymbol{h}|^2+\frac{\sigma^2}{|\boldsymbol{y}|^2}} \cdot \left( \boldsymbol{y}+ \sigma \cdot \boldsymbol{h} \boldsymbol{\epsilon} \right) \\
& =  \frac{|\boldsymbol{h}|^2}{\sqrt{\bar{\alpha}_{u}}\left(|\boldsymbol{h}|^2+\frac{\sigma^2}{|\boldsymbol{y}|^2}\right)} \cdot \left(\sqrt{\bar{\alpha}_{u}} \cdot \boldsymbol{y}_0 \right.\\
&~~~~~~~ \left. + \sqrt{1-\bar{\alpha}_{u}} \cdot  \frac{\sqrt{\bar{\alpha}_{u}}\sigma \cdot \boldsymbol{h}}{\sqrt{1-\bar{\alpha}_{u}} } \boldsymbol{\epsilon}\right).
\label{fading}
\end{aligned}
\end{equation}

Based on the Eqn. (\ref{fading}), when $|\frac{\sqrt{\bar{\alpha}_u} \sigma \cdot \boldsymbol{h}}{\sqrt{1-\bar{\alpha}_u}}| = 1$, i.e., $\bar{\alpha}_{u}= \frac{1}{1 + |\sigma \cdot\boldsymbol{h}|^2}$, the transmission process over the Rayleigh channel is equivalent to the $u$-th step forward process in the diffusion process. This equivalence can be expressed as:
\begin{equation}
\left. p_c\left(\boldsymbol{s}_{\text{MSE}} \mid \boldsymbol{y} \right) \sim p\left(\boldsymbol{y}_{u} \mid \boldsymbol{y}_0 \right) \right|_{\bar{\alpha}_{u}= \frac{1}{1 + |\sigma \cdot\boldsymbol{h}|^2}}. 
\end{equation}

For the MIMO channel, we utilize SVD decomposition \cite{10597355} to decouple the multi-antenna signals into independent streams, ensuring that the transformed signals approximate the Gaussian-distributed features required by the diffusion model. Given the channel state information (CSI) \cite{10660530,Deep2024Guo}, the MIMO channel matrix $\mathbf{H} \in \mathbb{C}^{M \times M}$ is decomposed as follows:
\begin{equation}
\boldsymbol{H}=\boldsymbol{U} \boldsymbol{\Sigma}_{\boldsymbol{H}} \boldsymbol{V}^H,
\end{equation}
where $\mathbf{U} \in \mathbb{C}^{M \times M}$ is a unitary matrix that satisfies $\mathbf{U} \mathbf{U}^H=\mathbf{I}$, representing the left singular vectors of $\mathbf{H}$. Similarly, $\mathbf{V} \in \mathbb{C}^{M \times M}$ is also a unitary matrix, satisfying $\mathbf{V} \mathbf{V}^H=\mathbf{I}$, and it represents the right singular vectors of $\mathbf{H}$. $\mathbf{V}^H$ is the Hermitian transpose of $\mathbf{V}$. $\boldsymbol{\Sigma}_{\boldsymbol{H}} \in \mathbb{R}^{M \times M}$ is a diagonal matrix, whose entries are the singular values of $\mathbf{H}$, $\sigma_1, \sigma_2, \ldots, \sigma_M$, arranged in descending order. This is expressed as:
\begin{equation}
\boldsymbol{\Sigma}_{\boldsymbol{H}}=\operatorname{diag}\left(\sigma_1, \sigma_2, \ldots, \sigma_M\right), \quad \sigma_1 \geq \sigma_2 \geq \cdots \geq \sigma_M \geq 0.
\end{equation}
The transmitted signal $\boldsymbol{x}_{\text{MIMO}}$ over MIMO channel can be represented as:
\begin{equation}
\boldsymbol{x}_{\text{MIMO}}=\mathbf{V} \boldsymbol{y}.
\end{equation}
The received signal $\boldsymbol{y}_{\text{MIMO}}$ is given as:
\begin{equation}
\boldsymbol{y}_{\text{MIMO}} = \boldsymbol{H} \boldsymbol{x}_{\text{MIMO}} + \sigma \cdot \boldsymbol{\epsilon}.
\end{equation}

By applying the unitary matrix $\mathbf{U}^H$ to the received signal $\boldsymbol{y}_{\text{MIMO}}$, the signal is transformed as:
\begin{equation}
\begin{aligned}
\boldsymbol{y}_{\text{MIMO}}^{\prime}&=\mathbf{U}^H \boldsymbol{y}_{\text{MIMO}}=\mathbf{U}^H\left(\mathbf{U} \boldsymbol{\Sigma}_{\boldsymbol{H}} \mathbf{V}^H \mathbf{V} \boldsymbol{y}+ \sigma \cdot \boldsymbol{\epsilon}\right) \\
&=\boldsymbol{\Sigma}_{\boldsymbol{H}} \boldsymbol{y}+ \sigma \cdot \boldsymbol{\epsilon}^{\prime},
\end{aligned}
\end{equation}
where $\boldsymbol{\epsilon}^{\prime} = \mathbf{U}^H \boldsymbol{\epsilon} $ is the transformed noise, which remains Gaussian due to the unitary property of $\mathbf{U}$. This separates the $M$-dimensional MIMO channel into $M$ independent scalar sub-channels, represented as:
\begin{equation}
\begin{aligned}
\boldsymbol{y}_i^{\prime} & =\sigma_i \cdot \boldsymbol{y}_i + \sigma \cdot \boldsymbol{\epsilon}_i^{\prime} \\
& =  \frac{\sigma_i}{\sqrt{\bar{\alpha}_{u}}} \cdot \left(\sqrt{\bar{\alpha}_{u}} \cdot \boldsymbol{y}_i + \frac{\sqrt{\bar{\alpha}_{u}}\sigma}{\sqrt{1-\bar{\alpha}_{u}} \sigma_i } \sqrt{1-\bar{\alpha}_{u}} \cdot \boldsymbol{\epsilon}_i^{\prime}\right),
\end{aligned}
\end{equation}
where $\sigma_i$ is the gain of the $i$-th sub-channel, $i=1,2, \ldots, M$. $\boldsymbol{y}_i^{\prime}$ represents the output of the $i$-th sub-channel after transformation. $\boldsymbol{y}_i$ represents the transmitted signal over the $i$-th sub-channel. $\boldsymbol{\epsilon}_i^{\prime}$ represents the noise component for the $i$-th sub-channel. When $|\frac{\sqrt{\bar{\alpha}_u} \sigma}{\sqrt{1-\bar{\alpha}_u} \sigma_i}| = 1$, i.e., $\bar{\alpha}_{u}= \frac{1}{1+\left(\sigma/\sigma_i\right)^2}$, the transmission process over the sub-channel of MIMO is equivalent to the $u$-th step forward process in the diffusion process. This equivalence can be expressed as:
\begin{equation}
\left. p_c\left( \boldsymbol{y}_i^{\prime} \mid \boldsymbol{y}_i \right) \sim p\left(\boldsymbol{y}_{u} \mid \boldsymbol{y}_0 \right) \right|_{\bar{\alpha}_{u}= \frac{1}{1+\left(\sigma/\sigma_i\right)^2}}.
\end{equation}

\section{The Details of the Diffusion-Driven Semantic Communication System with VAE-Based Compression}
\begin{figure*}
\centering
\includegraphics[width=\linewidth]{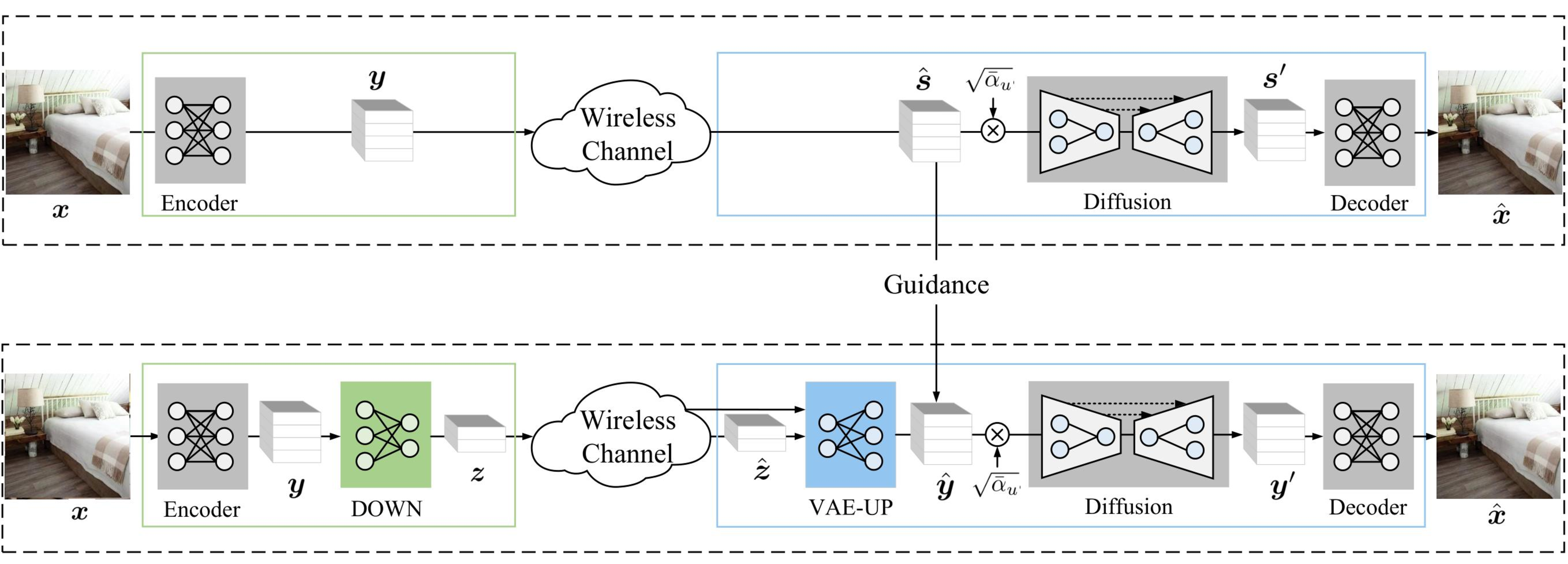}
\caption{The details of the proposed diffusion-driven semantic communication with VAE-based compression. In the gray boxes, we utilize existing parameters without the model training.}
\label{diffusion}
\end{figure*}

In this section, we present a diffusion-driven semantic communication framework enhanced by VAE-based compression. To ensure the Gaussian distribution of features, we employ VAE-based reparameterization for feature reconstruction. Additionally, we introduce guidance to focus on training the paired upsampling and downsampling networks to reduce training costs. The pre-trained diffusion model integrated with compression leverages guidance from the distribution of the pre-trained model, enabling the efficient generation of corresponding images with reduced bandwidth. Fig. \ref{diffusion} illustrates the framework for detailed semantic communication with compression and guidance.

\subsection{Compression and Reconstruction by Reparameterization}
\begin{figure}
\centering
\includegraphics[width=\linewidth]{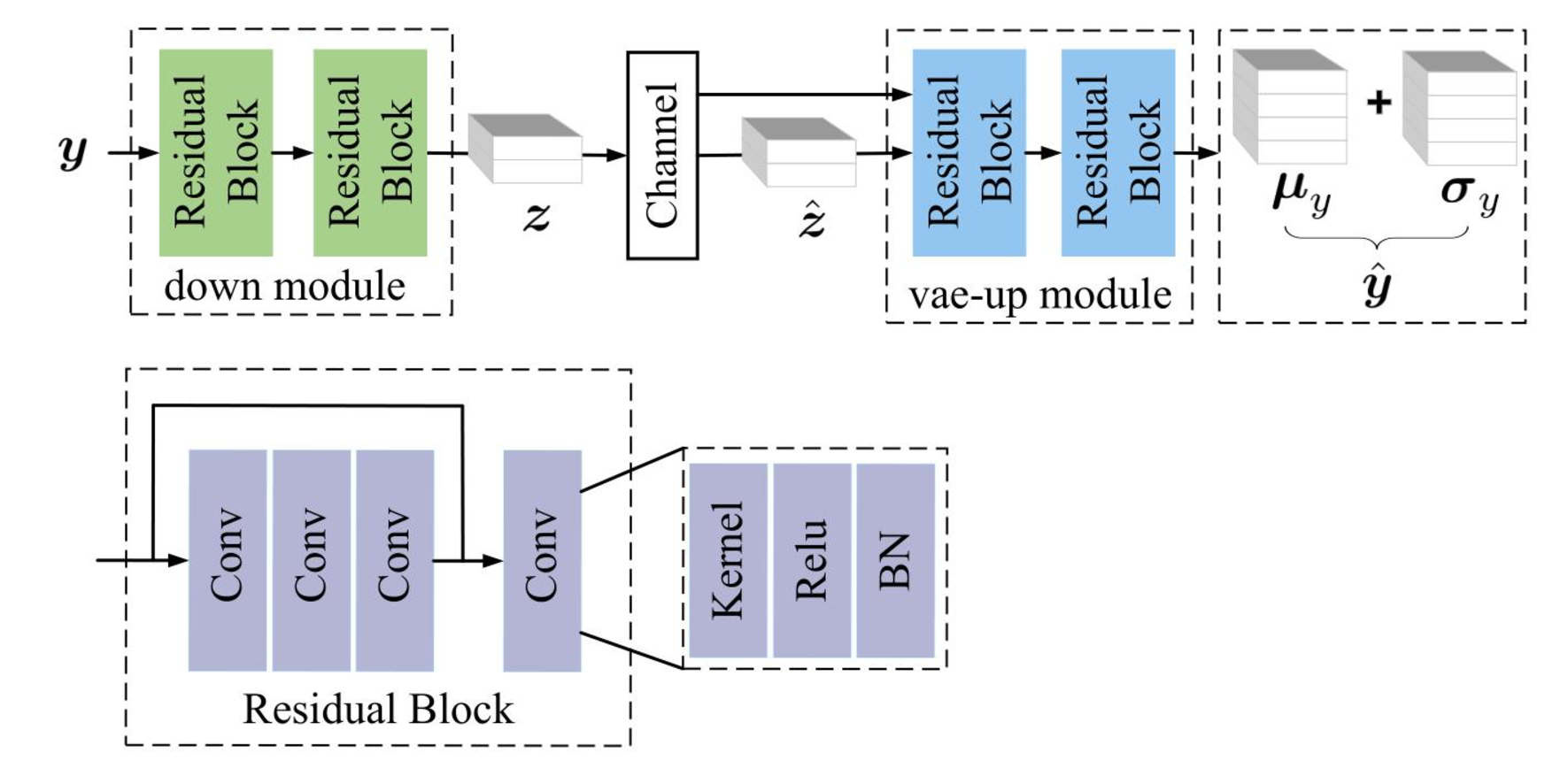}
\caption{The sturcture of the proposed compression and reconstruction based on reparameterization.}
\label{compression}
\end{figure}

In this section, we introduce the details of the VAE-based compression module aimed at mitigating bandwidth requirements by reducing data volume at the transmitter. This is complemented by a corresponding reconstruction module at the receiver to recover information lost during compression. To address potential information loss caused by non-linear operations in the compression and upsampling modules, we propose a novel compression and reconstruction method through reparameterization. This method helps maintain the Gaussian distribution of the latent feature, as shown in Fig. \ref{compression}. This approach ensures that the output maintains a Gaussian distribution. The upsampling module, based on VAE \cite{kingma2013auto}, is employed to reconstruct the transmitted information while preserving its distribution.

For the compression module, acting as a downsampling network, the semantic feature $\boldsymbol{z}$ with reduced data volume is transmitted. To avoid the computational burden caused by complex network, we design a low-complexity downsampling network consisting of two residual blocks to compress the semantic information. After transmission through the wireless channel, the received semantic feature is reconstructed by an upsampling network based on VAE to produce the mean features $\boldsymbol{\mu}_{y}$ and variance features $\boldsymbol{\sigma}_{y}$ of $\hat{\boldsymbol{y}}$. Since the variance $\boldsymbol{\sigma}_{y}$ is particularly affected by the noise in the channel, we incorporate the SNR of the channel as an input to the output variance network. This integration of SNR enables the network to dynamically adjust the output variance based on the noise level in the channel, thereby improving its adaptability to diverse channel conditions. Consequently, this approach enhances the fidelity of semantic reconstruction and strengthens the resilience of system to noise across varying channel conditions. The reconstruction module can be represented as:
\begin{equation}
\hat{\boldsymbol{y}} = F_{u}(\hat{\boldsymbol{z}}, \text{SNR}; \boldsymbol{\omega}),
\end{equation}

The upsampling network with parameter $\boldsymbol{\omega}$ is also designed to consist of two residual blocks to avoid the computational load. The reconstructed feature $\hat{\boldsymbol{y}}$ is expected to have a similar distribution to the feature $\hat{\boldsymbol{s}}$ of the generator without compression, to reduce bandwidth requirements while maintaining performance.

\subsection{Training with Guidance}
To learn the distribution of the uncompressed generator, we employ a method similar to distillation to ensure that the reconstructed feature closely aligns with that of the uncompressed generator. The output distribution of the $\hat{\boldsymbol{s}}$ serves as the target distribution to guide the $\hat{\boldsymbol{y}}$ of the compressed generator, making the output distribution of the $\hat{\boldsymbol{y}}$ as close as possible to minimize performance loss due to bandwidth compression. This alignment is designed through the guided objective $\mathcal{L}_{g}$, which minimizes the KL divergence between $\hat{\boldsymbol{y}}$ and $\hat{\boldsymbol{s}}$. By effectively learning from the distribution of $\hat{\boldsymbol{s}}$, the compressed generator can learn its representations accordingly. $\mathcal{L}_{g}$ ensures efficient compression while preserving the essential the feature distribution of $\hat{\boldsymbol{s}}$, which can be represented as:
\begin{equation}
\mathcal{L}_{g} = D_{\text{KL}} \left(p_c\left(\hat{\boldsymbol{s}}\mid\boldsymbol{y}\right) \| p_{\boldsymbol{\omega}}\left(\hat{\boldsymbol{y}}\mid\hat{\boldsymbol{z}}\right)\right),
\label{lg} 
\end{equation}
where $D_{\text{KL}}\left(\cdot \right)$ denotes the KL divergence between the two distributions. $p_{\boldsymbol{\omega}}\left(\hat{\boldsymbol{y}}\mid\hat{\boldsymbol{z}}\right)$ denotes the conditional probability from the upsampling module based on reparameterization. 

\emph{Corollary 1:} The KL divergence between two Gaussian distributions, $p_c\left(\hat{\boldsymbol{s}}\mid \boldsymbol{y}\right) \sim \mathcal{N}\left(\boldsymbol{y}, \sigma^2 \cdot \boldsymbol{1}\right)$ and $p_{\boldsymbol{\omega}}\left(\hat{\boldsymbol{y}}\mid\hat{\boldsymbol{z}}\right) \sim \mathcal{N}\left(\boldsymbol{\mu}_{y}, \boldsymbol{\sigma}_{y}^2\right)$, can be expressed in terms of $\boldsymbol{y}, \sigma^2 \cdot \boldsymbol{1}, \boldsymbol{\mu}_{y}$ and $\boldsymbol{\sigma}_{y}^2$ as:
\begin{equation}
\mathcal{L}_{g} = \log\!\left(\frac{\boldsymbol{\sigma}_{y}}{\sigma \cdot \boldsymbol{1}}\right)\!+\!\frac{\sigma^2 \cdot \boldsymbol{1}\!+\! \left(\boldsymbol{\mu}_{y}\!-\!\boldsymbol{y}\right)^2}{2 \boldsymbol{\sigma}^2_{y}}\!-\!\frac{1}{2},
\end{equation}
where $\boldsymbol{\sigma}^2_{y}\!=\!\boldsymbol{\sigma}_{y} \otimes \boldsymbol{\sigma}_{y}$, $\left(\boldsymbol{\mu}_{y}\!-\!\boldsymbol{y}\right)^2\!=\!\left(\boldsymbol{\mu}_{y}\!-\!\boldsymbol{y}\right) \otimes \left(\boldsymbol{\mu}_{y}\!-\!\boldsymbol{y}\right)$.
\emph{Proof:} The proof is given in Appendix A.

This guidance enables the compressed generator to learn from the uncompressed generator in bandwidth-constrained scenarios, utilizing the diffusion capabilities of semantic extraction. For instance, the guidance is used to train the compressed generator to transmit semantic features at lower bandwidth while retaining the generation capabilities of the Stable Diffusion. Moreover, this guidance helps to focus training efforts primarily on the supplementary downsampling and upsampling networks, concentrating on the inputs and outputs of these components while freezing the other structure of the generator. By focusing on these specific components, the overall training process is streamlined, reducing computational expenses while enhancing the performance of the compressed generator through learning from the distribution of the uncompressed generator.

\subsection{The Hybrid Loss from Reparameterization and Guidance}
In the described system, the hybrid loss function is composed of two key components. The first component is the modeling loss, derived from the reparameterization of the VAE. This loss measures the discrepancy between the input data and its reconstruction by the VAE, considering the stochastic nature of the latent space. The second component is the guidance loss, as mentioned earlier. It aims to align the distribution of the compressed generator with that of the uncompressed generator. This loss encourages the compressed generator to learn from the distribution of the uncompressed generator, with the objective of mitigating errors introduced due to bandwidth compression during the transmission. The overall loss function for the compression diffusion model within guidance can be expressed as follows:
\begin{equation}
\mathcal{L}(\boldsymbol{\psi}, \boldsymbol{\omega}, \boldsymbol{\phi}) = \mathcal{L}_{v} + \gamma \mathcal{L}_{g},
\end{equation}
where $\mathcal{L}_{v}$ represents the modeling loss function based on VAE, and $\gamma$ is a hyperparameter that controls the balance between the two components. 

Additionally, we regard the supplementary downsampling and upsampling networks before reparameterization as the VAE-based encoder. Since both the reverse process of diffusion and the VAE-based decoder offer posterior probability distributions, the reverse process of diffusion is utilized as the VAE-based decoder in this system. According to the VAE, the loss function $\mathcal{L}_{v}$ can be expressed as:
\begin{equation}
\mathcal{L}_{v}=\underbrace{\lambda D_{\text{KL}}\left(q_{\boldsymbol{\psi},\boldsymbol{\omega}}\left(\hat{\boldsymbol{y}}\mid\!\boldsymbol{y}\right)\|p_{\boldsymbol{\phi}}\left(\hat{\boldsymbol{y}}\right)\right)}_{\mathcal{L}_{\text{KL}}}+\underbrace{\mathbb{E}\!\left[\left({\boldsymbol{y}^{\prime}\!-\! \boldsymbol{y}}\right)^2\!\right]}_{\mathcal{L}_{\text{MSE}}},
\end{equation}
where $q_{\boldsymbol{\psi},\boldsymbol{\omega}}\left(\hat{\boldsymbol{y}} \mid \boldsymbol{y}\right)$ represents a prior probability provided by the supplementary downsampling and upsampling networks acting as the VAE-based encoder. The distribution $p_{\boldsymbol{\phi}}(\hat{\boldsymbol{y}})$ follows a standard Gaussian distribution $\mathcal{N}(\boldsymbol{0}, \boldsymbol{1})$ as \cite{kingma2013auto}. The hyperparameter $\lambda$ adjusts the balance between the first and second terms of the loss function $\mathcal{L}_{v}$. 

\emph{Corollary 2:} The first term of the loss function $\mathcal{L}_{v}$, which is the KL divergence between two Gaussian distributions, $q_{\boldsymbol{\psi},\boldsymbol{\omega}}\left(\hat{\boldsymbol{y}} \mid \boldsymbol{y} \right) \sim \mathcal{N}\left(\boldsymbol{\mu}_{y}, \boldsymbol{\sigma}_{y}^2 \right)$ and $p_{\boldsymbol{\phi}}(\hat{\boldsymbol{y}}) \sim \mathcal{N}(\boldsymbol{0}, \boldsymbol{1})$, can be expressed as:
\begin{equation}
D_{\text{KL}}\left(q_{\boldsymbol{\psi},\boldsymbol{\omega}}\left(\hat{\boldsymbol{y}}\mid\boldsymbol{y}\right)\|p_{\boldsymbol{\phi}}\left(\hat{\boldsymbol{y}}\right)\right)=\frac{1}{2}\left(\boldsymbol{\mu}^2_{y}+\boldsymbol{\sigma}^2_{y}-\log \boldsymbol{\sigma}^2_{y}-\boldsymbol{1}\right).
\end{equation}
\emph{Proof:} The proof is along the lines of the proof in Appendix A and is therefore omitted.

\emph{Corollary 3:} $\mathbb{E}\left[\left({\boldsymbol{y}^{\prime}-\boldsymbol{y}}\right)^2\right]$, which is the second term of the loss function $\mathcal{L}_{v}$, can be expressed as:
\begin{equation}
\mathbb{E}\left[\left({\boldsymbol{y}^{\prime}  - \boldsymbol{y}} \right)^2\right] = \mathbb{E}\left[\left(\hat{\boldsymbol{y}} - \hat{\boldsymbol{s}}\right)^2\right] + C.
\end{equation}
\emph{Proof:} The proof is given in Appendix B.

%Since $\hat{\boldsymbol{z}}_{st} \sim \mathcal{N}(\boldsymbol{\mu}_{st}, \boldsymbol{\sigma}^2_{st})$, $\hat{\boldsymbol{z}}_{te} \sim \mathcal{N}(\sqrt{\bar{a}}_{u} \boldsymbol{z}_{te}, 1-\bar{a}_{u}\boldsymbol{I})$, Eqn. \ref{loss_function} can be expressed as:
%\begin{equation}
%\begin{aligned}
%&\mathbb{E}\left[\left({\boldsymbol{z}_{st}^{\prime}  - \boldsymbol{z}_{te}} \right)^2\right] \\
%= & \mathbb{E}\left[\frac{1}{\bar{a}_{u}}\left(\hat{\boldsymbol{z}}_{te} - \hat{\boldsymbol{z}}_{st}\right)^2\right] \\
%= & \frac{1}{\bar{a}_{u}} \left[ \left(\boldsymbol{\mu}_{st} - \sqrt{\bar{a}}_{u} \boldsymbol{z}_{te} \right)^2 + \boldsymbol{\sigma}^2_{st} + (1-\bar{a}_{u}) \boldsymbol{I} \right]
%\end{aligned}
%\end{equation}

%The loss function based on VAE $\mathcal{L}_{v}$ can be rewritten as:
%\begin{equation}
%\begin{aligned}
%\mathcal{L}_{v} = - & \frac{{\lambda\bar{a}_{u}}}{2}\left(\log \boldsymbol{\sigma}^2_{st}-\boldsymbol{\mu}^2_{st}-\boldsymbol{\sigma}^2_{st}+\boldsymbol{1}\right) \\
%& + \left(\boldsymbol{\mu}_{st} - \sqrt{\bar{a}}_{u} \boldsymbol{z}_{te} \right)^2 + \boldsymbol{\sigma}^2_{st} + C
%\end{aligned}
%\end{equation}

In summary, the hybrid loss function of our system can be represented as:
\begin{equation}
\begin{aligned}
&\mathcal{L}(\boldsymbol{\psi}, \boldsymbol{\omega}, \boldsymbol{\phi})\\
%=&  \mathcal{L}_{v} + \gamma \mathcal{L}_{g} \\
=&\lambda D_{\text{KL}}\left(q_{\boldsymbol{\psi},\boldsymbol{\omega}}\left(\hat{\boldsymbol{y}}\mid\boldsymbol{y}\right)\|p_{\boldsymbol{\phi}}\left(\hat{\boldsymbol{y}}\right)\right) + \mathbb{E}\left[\left(\hat{\boldsymbol{y}} - \hat{\boldsymbol{s}}\right)^2\right] \\
&~~~ + \gamma D_{\text{KL}}\left(p_c\left(\hat{\boldsymbol{s}}\mid\boldsymbol{y}\right) \| q_{\boldsymbol{\psi},\boldsymbol{\omega}}\left(\hat{\boldsymbol{y}}\mid\boldsymbol{y}\right) \right) \\
=& \underbrace{\frac{\lambda}{2}\left(\boldsymbol{\mu}^2_{y}+\boldsymbol{\sigma}^2_{y}-\log\boldsymbol{\sigma}^2_{y}-\boldsymbol{1}\right) + \mathbb{E}\left[\left(\hat{\boldsymbol{y}} - \hat{\boldsymbol{s}}\right)^2\right]}_{\mathcal{L}_{v}} \\
&~~~ + \gamma \underbrace{\left(\log\!\left(\frac{\boldsymbol{\sigma}_{y}}{\sigma \cdot \boldsymbol{1}}\right)\!+\!\frac{\sigma^2 \cdot \boldsymbol{1}\!+\! \left(\boldsymbol{\mu}_{y}\!-\!\boldsymbol{y}\right)^2}{2 \boldsymbol{\sigma}^2_{y}}\right)}_{\mathcal{L}_{g}}.
%=& \left(\gamma- \lambda \bar{a}_{u} \right) \log \boldsymbol{\sigma}_{st} + \frac{\lambda \bar{a}_{u}+1}{2}\boldsymbol{\sigma}^2_{st} + \frac{\gamma(1- \bar{a}_{u})}{2\boldsymbol{\sigma}^2_{st}}\\
%& + \frac{{\lambda\bar{a}_{u}}}{2}\boldsymbol{\mu}^2_{st}+\!\left(\!\frac{\gamma}{2 \boldsymbol{\sigma}^2_{st}}\!+\!1\!\right)\!\left(\!\boldsymbol{\mu}_{st}\!-\!\sqrt{\bar{a}}_{u}\!\boldsymbol{z}_{te}\!\right)^2+ C
\end{aligned}
\end{equation}

This hybrid loss function integrates components that ensure compression and reconstruction. The term of $\mathcal{L}_{v}$ addresses the reconstruction through two primary components: the KL divergence between the distributions as well as the mean squared error between the reconstructed signal $\hat{\boldsymbol{y}}$ and the original received signal $\hat{\boldsymbol{s}}$. The term of $\mathcal{L}_{g}$ focuses on guiding the compression process by penalizing the KL divergence between the conditional distribution $p_c\left(\hat{\boldsymbol{s}}\mid\boldsymbol{y}\right)$ and $q_{\boldsymbol{\psi},\boldsymbol{\omega}}\left(\hat{\boldsymbol{y}}\mid\boldsymbol{y}\right)$. This ensures that the reconstructed distribution aligns closely with the target distribution, enhancing the semantic reconstruction.

\section{Results}
In this section, we provide a detailed description of the experimental setup and present experimental results to validate the effectiveness of the proposed semantic communication generation with compression and guidance.

\subsection{Experiment Setup}
\subsubsection{Dataset and Simulation Setting}
Our method was trained on the LSUN-Churches dataset, which contains 126000 images, and evaluated on both the LSUN-Churches and LSUN-Bedrooms datasets \cite{2015LSUN} with 300 images. The pre-trained diffusion model used in our approach is Stable Diffusion v2 \cite{9878449}. We configure the number of diffusion timesteps to $T=1000$ with a linear schedule, set the learning rate to $1 \times 10^{-4}$, and use the training batch size of 4. For the compression module and the VAE-based upsampling module, to reduce the added computational complexity, the residual block is a bottleneck structure composed of three units, which includes a convolutional filter followed by the LeakyReLU and batch normalization. The latent variables of the VAE are modeled to follow a Gaussian distribution through the reparameterization trick. To evaluate the effectiveness of our approach, we transform received signals into appropriate features suitable for the forward process of the diffusion model and extend the transformation to more complex wireless channels, including Rayleigh and MIMO channels. Specifically, we combine our approach with MSE equalization under the assumption that the channel gain $|\boldsymbol{h}|=1$ for the Rayleigh channel, while utilizing SVD decomposition \cite{10597355} for the MIMO channel with $M=2$. In addition, we also visualized the semantic distinctions between the generated images and real images. This helps to provide a more intuitive understanding of the effectiveness of generating semantic communication based diffusion. The simulations are conducted on the computer equipped with an Intel Xeon Silver 4110 CPU @ 2.10GHz and an NVIDIA RTX A40 GPU.
%The diffusion steps $T$ for the pre-trained model is 1000, the learning rate is $1×10^{-4}$ , the training batchsize is 16, and the noise schedule is linear with the increase . 300张测试图像 126,000训练图像
%, FID \cite{fid} 
\subsubsection{The Metrics}
The performance is evaluated in terms of details transmission and semantic transmission using multiple metrics such as peak signal-to-noise ratio (PSNR) and structural similarity index (SSIM) \cite{5596999}, learned perceptual image patch similarity (LPIPS) \cite{8578166}, Frechet Inception Distance (FID) \cite{fid} and CLIP-score \cite{2021Learning}. The quality of transmission is traditionally evaluated using distortion measures such as the PSNR and SSIM, which focus on pixel-level evaluation. PSNR emphasizes pixel-level errors, while SSIM considers the overall structural similarity of the pixel. Both metrics focus on evaluating the details of the images. 

In semantic communication, the evaluation of transmission quality revolves around how effectively the intended information is conveyed to the receiver. This necessitates a more comprehensive evaluation metric that considers not only pixel-level details but also the overall structure and content of the image.

%The learned perceptual image patch similarity (LPIPS) metric quantifies the similarity between two images based on human perception. It considers various factors such as texture, color, and structure, providing a more accurate assessment of perceptual quality compared to PSNR and SSIM. Additionally, metric CLIP-score is used to evaluate the quality of generated images. It measures semantic similarity using a pre-trained model, focusing on the semantic content of the images, which prioritizes semantic similarity. Thus, using both metrics together can provide a more comprehensive evaluation of the quality of generated images.

The LPIPS metric quantifies the similarity between two images based on human perception. It considers various factors such as texture, color, and structure, providing a more accurate assessment of perceptual quality compared to PSNR and SSIM. Additionally, metrics like FID and CLIP-score are used to evaluate the quality of generated images. FID measures the distance between the feature space distributions of generated and real images, focusing on global features. On the other hand, CLIP-score measures semantic similarity using a pre-trained model, focusing on the semantic content of the images. FID is more focused on the overall structure and texture of the image, while CLIP-score prioritizes semantic similarity. Thus, using both metrics together can provide a more comprehensive evaluation of the quality of generated images.

PSNR, SSIM, LPIPS, FID and CLIP-score are all employed as evaluation metrics in this study. The combination of these metrics allows for a more comprehensive evaluation of the performance of semantic communication systems in both details transmission and semantic transmission aspects. Higher values of PSNR, SSIM, and CLIP-score indicate better performance (indicated by $\uparrow$ in the following tables), whereas lower values of FID and LPIPS are preferred (indicated by $\downarrow$).
\subsection{Performance of Channel Integration and Adaptation}

\begin{table}[t]
\begin{center}
\caption{Comparison with different metrics in LSUN-Bedrooms datasets.}
\begin{tabular}{c|ccccc}
\hline
Metric & 200-steps & 3dB & 6dB & 9dB & 12dB \\
\hline
PSNR $\uparrow$ & 21.678 & -0.007 & -0.002 & +0.000 & +0.002 \\  %\cline{2-5}
\hline
SSIM $\uparrow$ & 0.652 & +0.000 & +0.000 & +0.000 & +0.000 \\  %\cline{2-5}
\hline
CLIP-score $\uparrow$ & 0.878 & +0.000 &	+0.000 & +0.000 & +0.000 \\
\hline
FID $\downarrow$ & 32.539 & +0.562 & 	+0.489 & +0.306 & +0.251 \\  %\cline{2-5}
\hline
LPIPS $\downarrow$ & 0.261 & +0.000 & +0.001 & +0.000 & -0.001 \\  %\cline{2-5}
\hline
\end{tabular}
\label{tab:bedroom}
\end{center}
\end{table}
\begin{table}[t]
\begin{center}
\caption{Comparison with different metrics in LSUN-Churches datasets.}
\begin{tabular}{c|ccccc}
\hline
Metric & 200-steps & 3dB & 6dB & 9dB & 12dB \\
\hline
PSNR $\uparrow$ & 19.379 & -0.001 & +0.002 & -0.003 & +0.001 \\  %\cline{2-5}
\hline
SSIM $\uparrow$ & 0.544 & +0.001 & +0.001 & -0.001 & +0.001 \\  %\cline{2-5}
\hline
CLIP-score $\uparrow$ & 0.878 & -0.004 &	+0.000 & -0.001 & +0.000 \\
\hline
FID $\downarrow$ & 30.678 & +0.611 & 	+0.587 & +0.533 & +0.388 \\  %\cline{2-5}
\hline
LPIPS $\downarrow$ & 0.273 & +0.000 & +0.001 & +0.000 & +0.000 \\  %\cline{2-5}
\hline
\end{tabular}
\label{tab:church}
\end{center}
\end{table}

\begin{table}[t]
\begin{center}
\caption{The adaptive generation in LSUN-Bedrooms datasets for different channels.}
\begin{tabular}{c|c|ccccc}
\hline
\multicolumn{1}{c|}{Metric} & Channel & 0dB & 3dB & 6dB & 9dB & 12dB \\
\hline
\multirow{3}{*}{PSNR $\uparrow$} & AWGN & 22.74 & 25.59 & 26.79 & 26.79 &	 27.83 \\
 & MIMO & 22.74 & 24.24 & 25.59 & 26.79 & 27.83 \\
 & Rayleigh & 14.87 & 17.70 & 21.23 & 23.27 & 24.39 \\
\hline
\multirow{3}{*}{SSIM $\uparrow$} & AWGN & 0.68 & 0.73 & 0.76 & 0.79 & 0.81 \\
 & MIMO & 0.68 & 0.73 & 0.76 & 0.79 & 0.81 \\
 & Rayleigh & 0.52 & 0.57 & 0.66 & 0.70 & 0.72 \\
\hline
\multirow{3}{*}{CLIP $\uparrow$} & AWGN & 0.88 &	0.90 & 0.91 & 0.91 & 0.91 \\
 & MIMO & 0.88 & 0.89 & 0.90 & 0.91 & 0.91 \\
 & Rayleigh & 0.53 & 0.66 & 0.82 & 0.83  & 0.85 \\
\hline
\multirow{3}{*}{FID $\downarrow$} & AWGN & 28.43 & 23.15 & 19.45 & 16.56 & 14.45 \\
 & MIMO & 28.58 & 23.41 & 19.38 & 16.52 & 14.31 \\
 & Rayleigh & 234.08 & 115.28 & 43.26 & 30.67 & 22.47 \\
\hline
\multirow{3}{*}{LPIPS $\downarrow$} & AWGN & 0.22 & 0.18 & 0.15 & 0.12 & 0.11 \\
 & MIMO & 0.22 & 0.18 & 0.15 & 0.12 & 0.11 \\
 & Rayleigh & 0.78 & 0.58 & 0.34 & 0.25 & 0.21 \\
\hline
\end{tabular}
\label{tab:adaptive_bedroom}
\end{center}
\end{table}

\begin{table}[t]
\begin{center}
\caption{The adaptive generation in LSUN-Churches datasets for different channels.}
\begin{tabular}{c|c|ccccc}
\hline
\multicolumn{1}{c|}{Metric} & Channel & 0dB & 3dB & 6dB & 9dB & 12dB \\
\hline
\multirow{3}{*}{PSNR $\uparrow$} & AWGN & 20.16 & 21.39 & 22.52 & 23.52 & 24.38 \\
 & MIMO & 20.16 & 21.39 & 22.52 & 23.52 & 24.38 \\
 & Rayleigh & 13.84 & 16.15 & 19.39 & 21.14 & 22.01 \\
\hline
\multirow{3}{*}{SSIM $\uparrow$} & AWGN & 0.58 & 0.62 & 0.67 & 0.70 & 0.73 \\
 & MIMO & 0.58 & 0.62 & 0.67 & 0.70 & 0.73 \\
 & Rayleigh & 0.41 & 0.45 & 0.55 & 0.60 & 0.63 \\
\hline
\multirow{3}{*}{CLIP $\uparrow$} & AWGN & 0.86 & 0.88 &	0.89 & 0.89 & 0.89 \\
 & MIMO & 0.86 & 0.88 & 0.89 & 0.89 & 0.89 \\
 & Rayleigh & 0.45 & 0.56 & 0.77 & 0.82 & 0.83 \\
\hline
\multirow{3}{*}{FID $\downarrow$} & AWGN & 27.17 & 22.43 & 19.12 & 16.52 & 14.72 \\
 & MIMO & 27.10 & 22.59 & 19.05 & 16.63 & 14.77 \\
 & Rayleigh & 272.50 & 151.58 & 43.80 & 37.96 & 29.50 \\
\hline
\multirow{3}{*}{LPIPS $\downarrow$} & AWGN & 0.24 & 0.20 & 0.17 & 0.14 & 0.13 \\
 & MIMO & 0.24 & 0.20 & 0.17 & 0.14 & 0.12 \\
 & Rayleigh & 0.86 & 0.70 & 0.42 & 0.28 & 0.23 \\
\hline
\end{tabular}
\label{tab:adaptive_church}
\end{center}
\end{table}

\begin{figure*}
\centering
\includegraphics[width=0.8\linewidth]{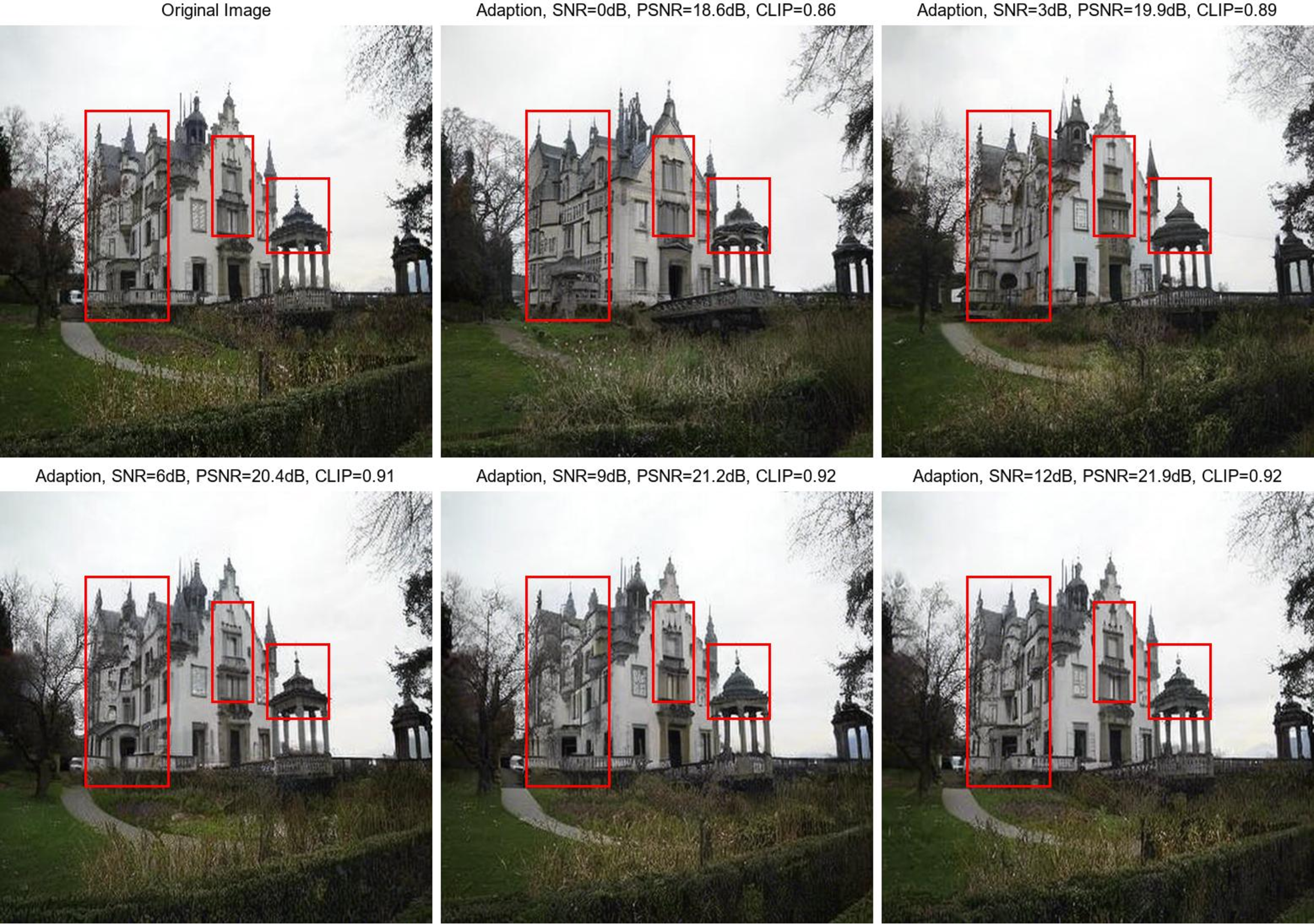}
\caption{The visual results of the adaptation generation across various SNR levels. When the SNR is low, the semantic information of the image can be correctly transmitted, and as the SNR increases, more details can be recovered. Noticeably different regions, highlighted with red boxes, are used for clear comparison to demonstrate how improved SNR contributes to more detailed reconstruction.}
\label{results}
\end{figure*}

The noise from the wireless channel is integrated into the forward process to mitigate its impact. At the receiver, the compensated noise is taken into account, utilizing the channel noise to complete the fixed 200 steps of the forward process. Tables \ref{tab:bedroom} and \ref{tab:church} present the results of compensating for additional noise up to 200 steps at the receiver under different SNR conditions in the LSUN-Bedrooms and LSUN-Churches datasets, respectively. The various metrics indicate that the generation results obtained by incorporating channel noise as part of the forward process are similar to those obtained with a fixed number of steps. For PSNR, SSIM, LPIPS, and CLIP-score, the maximum deviation does not exceed 0.01, while for FID, the maximum deviation does not exceed 1. This suggests that integrating the channel into diffusion-driven semantic communication eliminates the influence of the channel on the system.

We also evaluate the generation performance on the LSUN Bedrooms and LSUN Churches datasets, where the noise from the wireless channel is treated as a complete forward process to adapt to different SNR conditions. To further validate our approach, we extend the transformation to more complex wireless channels, such as Rayleigh and MIMO channels. The received signals are converted into appropriate features suitable for the forward process of the diffusion model. Specifically, we combine our approach with MSE equalization for the Rayleigh channel and utilize SVD decomposition for the MIMO channel. As shown in Tables \ref{tab:adaptive_bedroom} and \ref{tab:adaptive_church}, the results demonstrate that the performance of the MIMO channel, after applying SVD decomposition, closely approximates that of the AWGN channel. At high SNR levels, the Rayleigh channel, when combined with MSE equalization, achieves performance comparable to the AWGN channel. These results further validate the adaptability and robustness of our approach across diverse wireless channels. The CLIP-score metrics, used for evaluating the semantic content of the images, demonstrate indicate that diffusion-driven semantic communication ensures accurate transmission of semantic information across various SNR levels. However, at higher SNR, diffusion-driven semantic communication not only ensures accurate transmission of semantic information but also better preserves the details of the images. This result is reflected in the pixel-level evaluation metrics such as PSNR and SSIM. The visual results of generated images at different SNR levels are illustrated in Fig. \ref{results}. Even at lower SNR, semantic information can be effectively transmitted. For instance, in the figure, when comparing SNR = 0 dB with SNR = 12 dB, differences (highlighted with red boxes) can be observed in the details of the church's roof, windows, and architectural style. At lower SNR, the overall semantic content remains consistent, but details are lost. As the SNR increases, these details are better recovered in images.

\subsection{Performance of Compression with Different Channels and Bandwidths}

\begin{figure*}
    \centering
    \subfigure[Comparison of PSNR metric.]{\label{4c}\includegraphics[width=0.24\hsize, height=0.2\hsize]{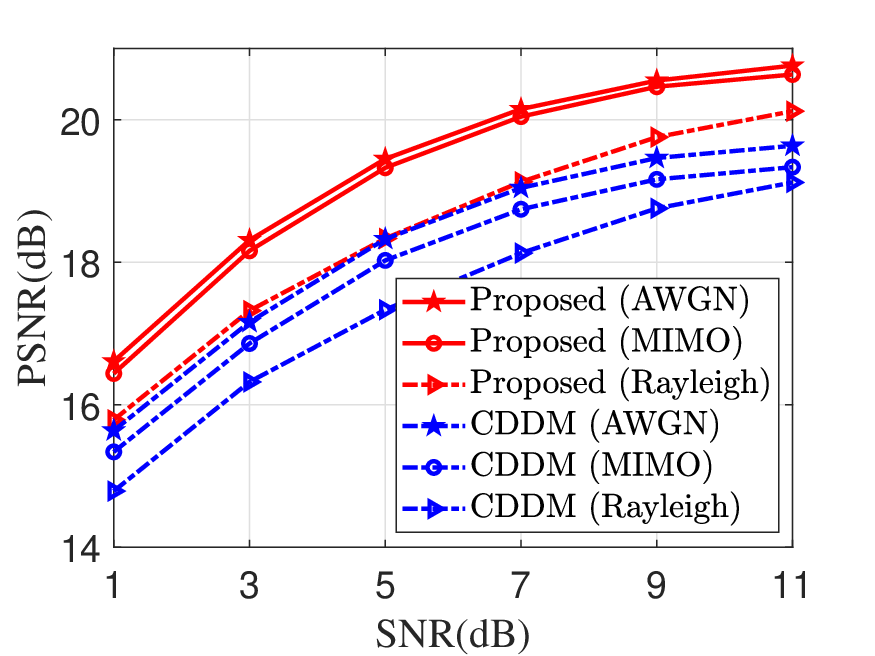}}
    \subfigure[Comparison of SSIM metric.]{\label{4d}\includegraphics[width=0.24\hsize, height=0.2\hsize]{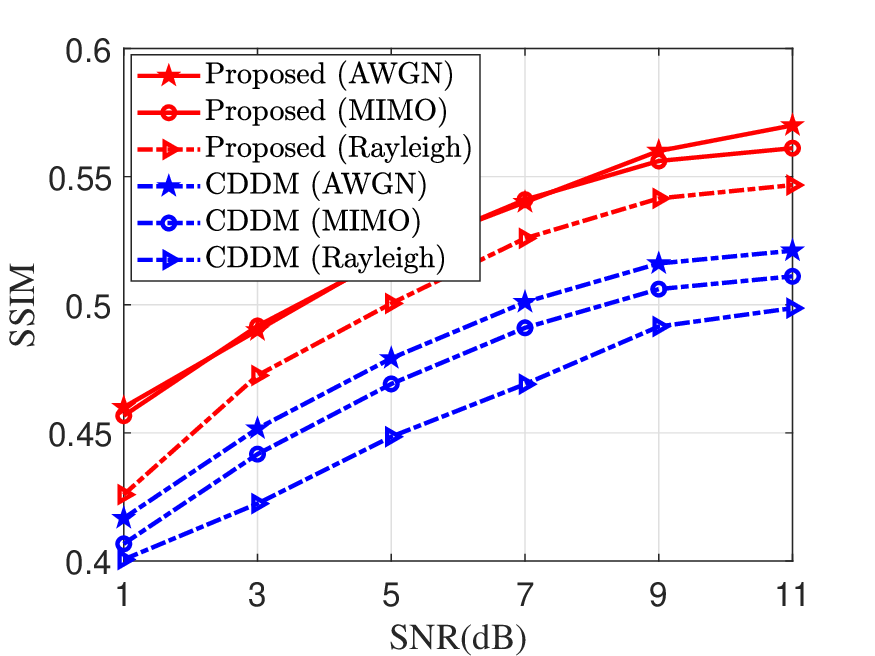}}
    \subfigure[Comparison of LIPIPS metric.]{\label{4f} \includegraphics[width=0.24\hsize, height=0.2\hsize]{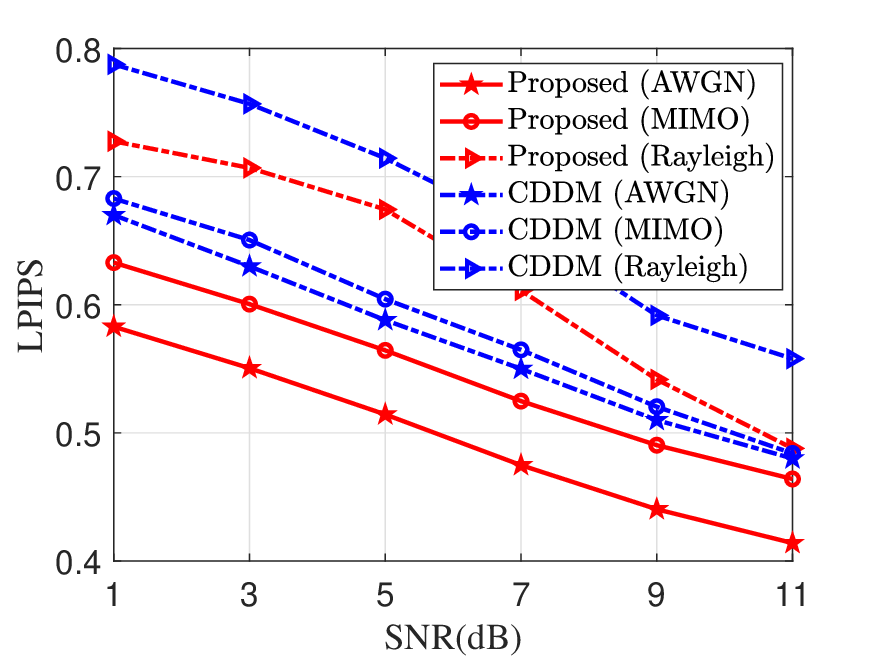}}
    \subfigure[Comparison of CLIP metric.]{\label{4e} \includegraphics[width=0.24\hsize, height=0.2\hsize]{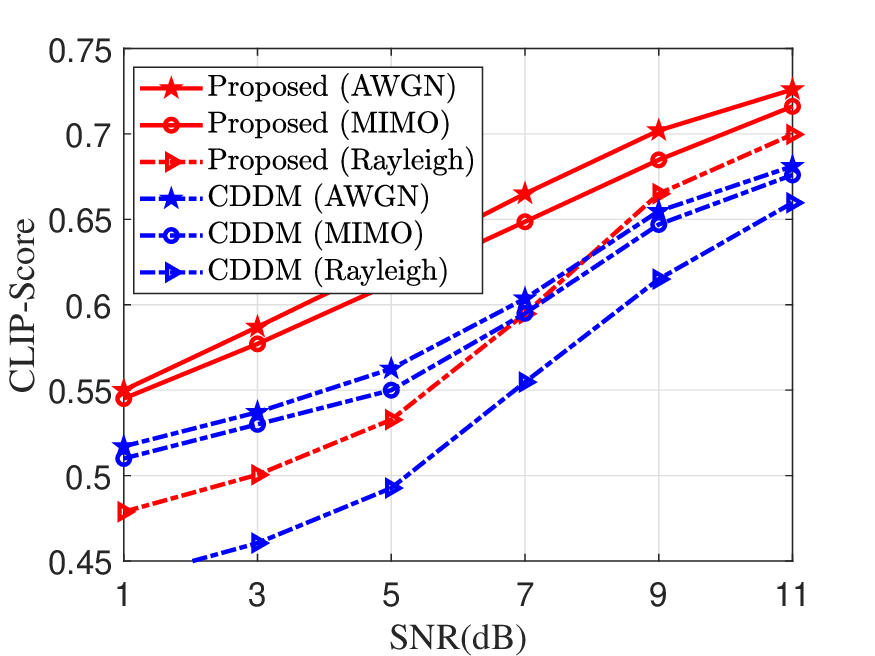}}
    \caption{The comparisons on LSUN-Churches dataset across different wireless channels.}
    \label{fig:summary_different_channels}
\end{figure*}

\begin{figure*}
    \centering
    \subfigure[Comparison of PSNR metric.]{\label{4c}\includegraphics[width=0.24\hsize, height=0.2\hsize]{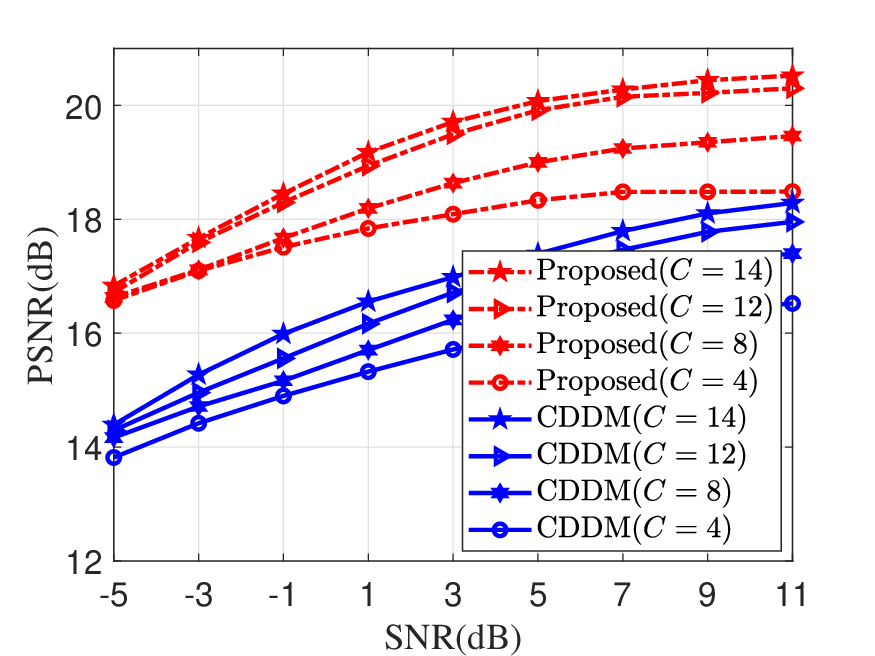}}
    \subfigure[Comparison of SSIM metric.]{\label{4d}\includegraphics[width=0.24\hsize, height=0.2\hsize]{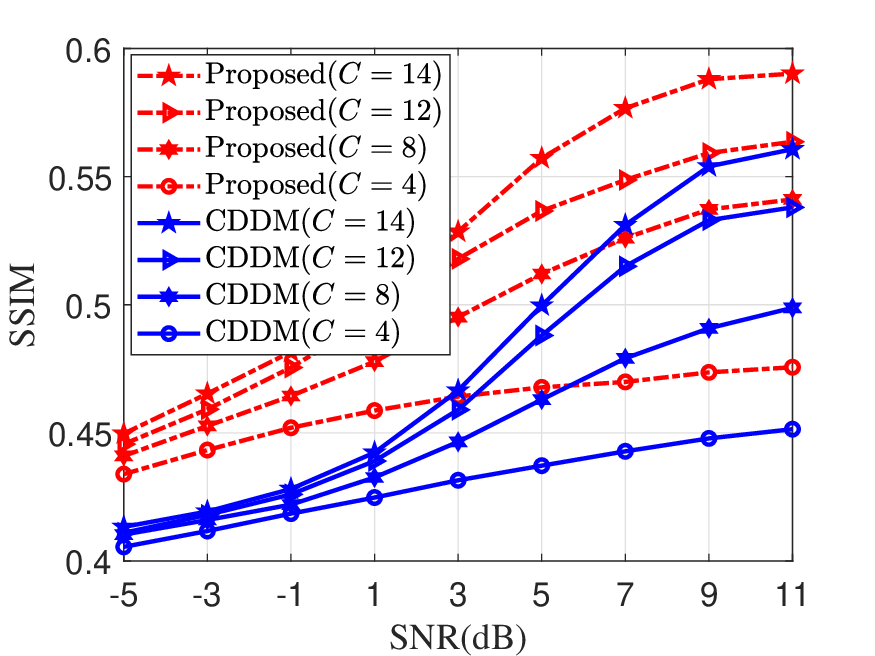}}
    \subfigure[Comparison of LIPIPS metric.]{\label{4f} \includegraphics[width=0.24\hsize, height=0.2\hsize]{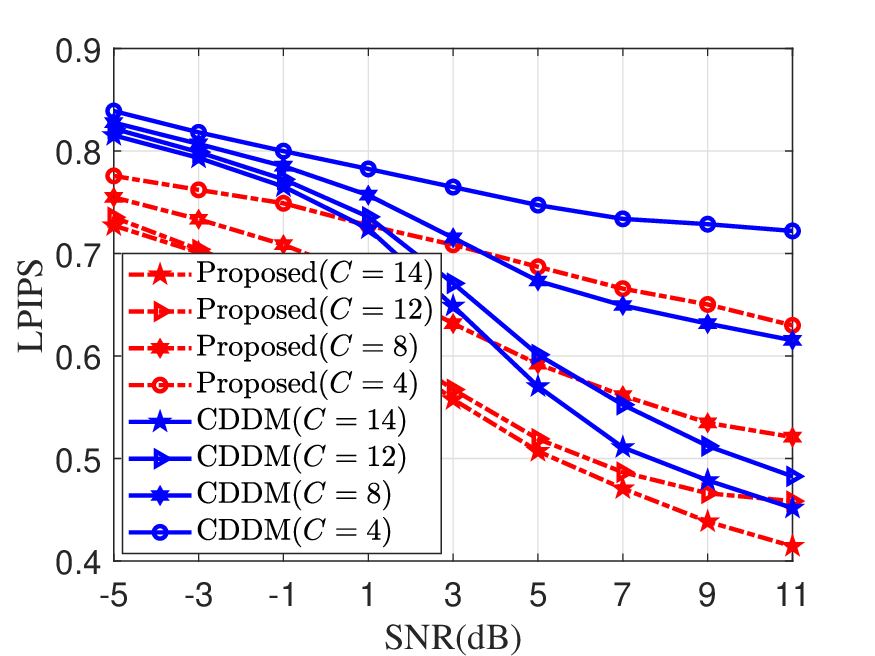}}
    \subfigure[Comparison of CLIP metric.]{\label{4e} \includegraphics[width=0.24\hsize, height=0.2\hsize]{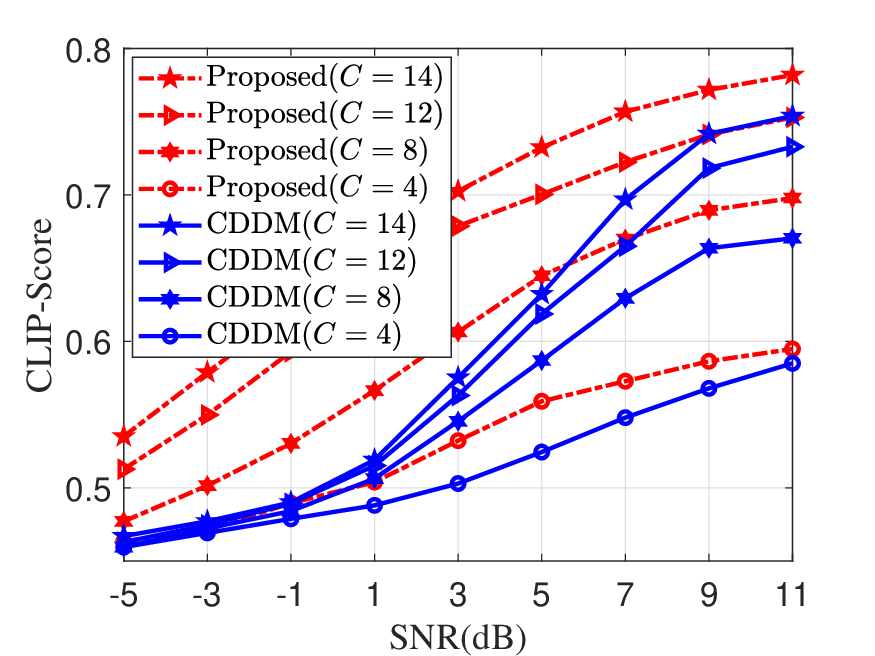}}
    \caption{The comparisons on LSUN-Churches dataset across different compression rates.}
    \label{fig:summary}
\end{figure*}

In this subsection, we evaluate the performance of the proposed generative semantic communication framework with compression and guidance across various compression rates. The method ``CDDM'' \cite{10480348}, which integrates diffusion models with the DeepJSCC network, is employed as our benchmark for performance comparison. While CDDM uses the same encoder and decoder network structures and incorporates guidance from the pre-trained diffusion model with the MSE loss function, our approach differs by introducing a VAE-based compression module for bandwidth adaptation and a hybrid loss function to further enhance reconstruction quality. Since previous studies have shown that DeepJSCC consistently outperforms classical separation-based methods, we exclude those methods from consideration as baselines.

\begin{table}
\centering
\caption{Experimental requirements of an epoch on the LSUN-Churches dataset.}
\begin{tabular}{c|c}
\hline
%\textbf{Metric} & \textbf{Requirement} \\ \hline
Training time (Memory usage) & 34 h (27139 MiB) \\ \hline
Inference time (Memory usage) & 7 s (26125 MiB) \\ \hline
\end{tabular}
\label{tab:hyperparameters}
\end{table}

We compare the performance of our method and CDDM under different channels. Additionally, we further evaluate and compare their performance under AWGN channels across various bandwidth constraints. Fig. \ref{fig:summary_different_channels} and Fig. \ref{fig:summary} present the comparisons in terms of PSNR, SSIM, LPIPS, and CLIP-score metrics, respectively. As shown in Fig. \ref{fig:summary_different_channels}, our proposed method achieves better performance across all evaluated channels. In the Fig. \ref{fig:summary}, the legends with ``Proposed ($C$)'' and ``CDDM ($C$)'' represent our proposed generation semantic communication system and the CDDM system with different compression rates, respectively. Here, $C$ denotes the channel dimensionality of the transmitted features, where fewer channels indicate higher compression rates. The mapping between compression rate $r$ and $C$ is defined as $r=0.0013C$. Our objective is to ensure accurate transmission of semantic information while using less bandwidth. Fig. \ref{fig:summary} shows the comparisons in terms of PSNR, SSIM, LPIPS and CLIP-score metrics, respectively. The results show that our proposed generation semantic communication system has significant performance improvement compared to the CDDM at all evaluated values of SNR and the compression rate. Further, Table \ref{tab:hyperparameters} presents the training and inference times, as well as the memory requirements of the proposed methods on the LSUN-Churches dataset. Specifically, the inference time is evaluated under 10 dB SNR, the reverse steps equivalently converted to 70.

\subsection{Ablation study}

\begin{table*}
\begin{center}
\caption{The ablation of the reparameterization and the proposed loss on LSUN-Churches dataset across different SNRs.}
\begin{tabular}{c|cccc|cccccc}
\hline
\multirow{2}{*}{Metric} & \multicolumn{4}{c|}{Methods} & \multicolumn{6}{c}{SNR} \\ 
\cline{2-11}
 & VAE & $L_{\text{MSE}}$ & $L_{\text{KL}}$ & $L_g$ & 1dB & 3dB & 5dB & 7dB & 9dB & 12dB \\ 
\hline
\multirow{6}{*}{PSNR $\uparrow$} &  & \checkmark & & & 13.79 & 14.39 & 14.56 & 14.82 & 14.96 & 15.02 \\
 & \checkmark & \checkmark & & & 16.15 & 17.51 & 18.32 & 18.72 & 18.94 & 19.09 \\
 & \checkmark &  &  & \checkmark & 15.78 & 17.06 & 17.73 & 17.92 & 17.95 & 18.02 \\
 & \checkmark & \checkmark & \checkmark & & 16.66 & 18.25 & 19.28 & 19.91 & 20.35 & 20.64 \\
 & \checkmark & \checkmark &  & \checkmark & 16.26 & 17.99 & 19.21 & 20.01 & 20.50 & 20.71 \\
 & \checkmark & \checkmark & \checkmark & \checkmark & 16.61 & 18.31 & 19.45 & 20.15 & 20.55 & \textbf{20.76} \\
\hline
\multirow{6}{*}{SSIM $\uparrow$} & & \checkmark & & & 0.35 & 0.35 & 0.35 & 0.36 & 0.37 & 0.37 \\
 & \checkmark & \checkmark & & & 0.43 & 0.45 & 0.47 & 0.48 & 0.49 & 0.49 \\
 & \checkmark &  &  & \checkmark & 0.40 & 0.43 & 0.44 & 0.44 & 0.45 & 0.45 \\
 & \checkmark & \checkmark & \checkmark & & 0.45 & 0.48 & 0.51 & 0.53 & 0.55 & 0.56 \\
 & \checkmark & \checkmark &  & \checkmark & 0.45 & 0.49 & 0.52 & 0.54 & 0.56 & 0.56 \\
 & \checkmark & \checkmark & \checkmark & \checkmark & 0.46 & 0.49 & 0.52 & 0.54 & 0.56 & \textbf{0.57} \\
\hline
\multirow{6}{*}{CLIP $\uparrow$} & & \checkmark & & & 0.45 & 0.48 & 0.49 & 0.50 & 0.53 & 0.54 \\
 & \checkmark & \checkmark & & & 0.50 & 0.51 & 0.52 & 0.53 & 0.56 & 0.58 \\
 & \checkmark &  &  & \checkmark & 0.55 & 0.58 & 0.59 & 0.62 & 0.66 & 0.70 \\
 & \checkmark & \checkmark & \checkmark & & 0.54 & 0.56 & 0.58 & 0.61 & 0.64 & 0.69 \\
 & \checkmark & \checkmark &  & \checkmark & 0.56 & 0.58 & 0.60 & 0.64 & 0.68 & 0.72 \\
 & \checkmark & \checkmark & \checkmark & \checkmark & 0.56 & 0.58 & 0.61 & 0.65 & 0.70 & \textbf{0.74} \\
\hline
\multirow{6}{*}{FID $\downarrow$} & & \checkmark & & & 268.89 & 254.21 & 244.21 & 217.34 & 206.88 & 193.05 \\
 & \checkmark & \checkmark & & & 249.67 & 238.40 & 230.23 & 217.34 & 197.17 & 168.68 \\
 & \checkmark &  &  & \checkmark & 207.39 & 182.78 & 156.71 & 122.33 & 91.50 & 71.78 \\
 & \checkmark & \checkmark & \checkmark & & 210.74 & 191.18 & 169.47 & 135.49 & 102.14 & 77.67 \\
 & \checkmark & \checkmark &  & \checkmark & 196.70 & 172.22 & 142.69 & 108.16 & 78.06 & 61.77 \\
 & \checkmark & \checkmark & \checkmark & \checkmark & 193.54 & 168.72 & 139.46 & 102.83 & 71.39 & \textbf{59.87} \\

\hline
\multirow{6}{*}{LPIPS $\downarrow$} & & \checkmark & & & 0.70 & 0.69 & 0.69 & 0.68 & 0.68 & 0.68 \\
 & \checkmark & \checkmark & & & 0.69 & 0.69 & 0.69 & 0.68 & 0.68 & 0.67 \\
 & \checkmark &  &  & \checkmark & 0.64 & 0.60 & 0.57 & 0.53 & 0.49 & 0.47 \\
 & \checkmark & \checkmark & \checkmark & & 0.64 & 0.62 & 0.60 & 0.56 & 0.53 & 0.50 \\
 & \checkmark & \checkmark &  & \checkmark & 0.63 & 0.60 & 0.56 & 0.51 & 0.48 & 0.45 \\
 & \checkmark & \checkmark & \checkmark & \checkmark & 0.57 & 0.53 & 0.47 & 0.45 & 0.43 & \textbf{0.42} \\
\hline
\end{tabular}
\label{tab:ablation}
\end{center}
\end{table*}

To comprehensively evaluate the contributions of various components in our proposed framework, ablation experiment is provided under different SNR conditions. The results are summarized in Table Table \ref{tab:ablation}, where each row represents the performance of the system using a different combination of components, including the VAE module and specific loss terms, such as $L_{\text{MSE}}$ and $L_{\text{KL}}$ of $L_v$ and $L_g$. Metrics such as PSNR, SSIM, CLIP-score, FID, and LPIPS are employed to evaluate both the pixel-level reconstruction fidelity and the semantic-level consistency of the generated results.

The baseline method, which employs only $L_{\text{MSE}}$ without the VAE module, demonstrates the weakest performance across all metrics, particularly under low SNR conditions. Incorporating the VAE module to maintain Gaussian-distributed features results in significant improvements in PSNR and SSIM, reflecting enhanced fidelity in the reconstructed images. The inclusion of $L_{\text{KL}}$ further boosts performance, as it contributes to better regularization and consistency in the latent space representation. Additionally, the integration of $L_g$ , which aligns the semantic content with the guidance provided by the pre-trained diffusion model, consistently enhances all evaluated metrics. Overall, the integration of all components ($L_{\text{MSE}}$, $L_{\text{KL}}$, $L_g$, and the VAE module) achieves the best performance, as evidenced by the highest PSNR (20.76), SSIM (0.57), and CLIP score (0.74), alongside the lowest FID (59.87) and LPIPS (0.42). These results demonstrate the effectiveness of the proposed framework in both semantic-level and pixel-level evaluations. Each component contributes to the overall system performance, and their combined integration leads to an improvement in both semantic preservation and detailed reconstruction quality under bandwidth constraints.

\subsection{Hyperparameters Settings in the Hybrid Loss}
\begin{table}
\begin{center}
\caption{The performance of different $\lambda$ in LSUN-Churches datasets, when $\gamma=0$ and SNR=5dB.}
\begin{tabular}{c|cccccc}
\hline
$\lambda$ & 1 & 0.1 & 0.01 & 0.001 & 0.0001 & 0.00001 \\
\hline
PSNR $\uparrow$ & 17.75 &	18.78 & 18.65 & 18.52 & 18.44 & 18.41 \\  %\cline{2-5}
\hline
SSIM $\uparrow$ & 0.43 & 0.49 & 0.48 & 0.47 & 0.47 & 0.47 \\  %\cline{2-5}
\hline
CLIP-score $\uparrow$ & 0.57 & 0.58 &	 0.56 & 0.55 & 0.54 & 0.54 \\
\hline
FID $\downarrow$ & 193.7 & 171.4 & 187.1 & 194.5 & 195.1 & 195.0 \\  %\cline{2-5}
\hline
LPIPS $\downarrow$ & 0.68 & 0.60 & 0.66 & 0.66 & 0.67 & 0.67 \\  %\cline{2-5}
\hline
\end{tabular}
\label{tab:hyperparameter1}
\end{center}
\end{table}

\begin{table}
\begin{center}
\caption{The performance of different $\gamma$ in LSUN-Churches datasets, when $\lambda=0.1$ and SNR=5dB.}
\begin{tabular}{c|cccccc}
\hline
$\gamma$ & 1 & 0.1 & 0.01 & 0.001 & 0.0001 & 0.00001 \\
\hline
PSNR $\uparrow$ & 18.50 &	19.11 & 19.07 & 19.06 & 19.01 & 19.03 \\  %\cline{2-5}
\hline
SSIM $\uparrow$ & 0.47 & 0.50 & 0.50 & 0.50 & 0.49 & 0.49 \\  %\cline{2-5}
\hline
CLIP-score $\uparrow$ & 0.58 & 0.63 &	 0.60 & 0.59 & 0.58 & 0.58 \\
\hline
FID $\downarrow$ & 187.8 & 108.8 & 142.0 & 155.5 & 157.5 & 157.8 \\  %\cline{2-5}
\hline
LPIPS $\downarrow$ & 0.50 & 0.60 & 0.63 & 0.64 & 0.64 & 0.64 \\  %\cline{2-5}
\hline
\end{tabular}
\label{tab:hyperparameter2}
\end{center}
\end{table}

Here, our focus lies on configuring the hyperparameters within the loss function for compression and guidance, which is employed to train the diffusion model. Hyperparameters, predetermined variables set before the learning process begins and remain constant during training, play a pivotal role in balancing the diverse components of the loss function and can significantly impact the capacity of model to learn from data. We investigate various hyperparameter configurations and their implications on performance. To conserve training resources, we subset the training dataset to include only 1000 images and evaluate it on a testing dataset consisting of 300 images. Training for 20 epochs allows us to identify the optimal combination based on selected metrics. Initially, we set the hyperparameter $\gamma$ to 0 to determine the optimal value of the hyperparameter $\lambda$. Subsequently, after identifying the best value for $\lambda$, we adjust $\gamma$ to select the optimal combination. As depicted in Tab. \ref{tab:hyperparameter1} and Tab. \ref{tab:hyperparameter2}, we ultimately selected the parameter combination of $\gamma=0.1$ and $\lambda=0.1$, which exhibited the best performance.

\section{Conclusion}
In this paper, we introduced a novel diffusion-driven semantic communication framework that incorporates advanced VAE-based compression for generative model in bandwidth-constrained environments. By leveraging the diffusion model, where the signal transmission process over the wireless channel serves as the forward diffusion process, our architecture successfully eliminates Gaussian noise from channel while conserving bandwidth, enhancing its applicability in wireless communication scenarios. We integrated a downsampling module and a corresponding VAE-based upsampling module with reparameterization to ensure the recovered features conform to the Gaussian distribution, a prerequisite for the diffusion model. Furthermore, we derived the loss function with the guidance for this system based on the design of reparameterization and compression. Experimental results demonstrate the effectiveness of our approach, with improvements observed in both compression rates and SNR compared to the CDDM baseline. Specifically, the reparameterization significantly enhances pixel-level performance, while guidance plays a crucial role in improving semantic transmission. This comprehensive approach demonstrates the potential for advancing semantic communication systems, offering enhanced reliability in generation performance.
\appendices
\section{Proof of Theorem 1}
The point in the feature map $\hat{\boldsymbol{s}}$ and $\hat{\boldsymbol{y}}$ are considered as a conditional probability, represented by $p_c\left(\hat{s}\mid y\right) \sim \mathcal{N}\left(y, \sigma^2 \right)$ and $p_{\omega}\left(\hat{y}\mid\hat{z}\right) \sim \mathcal{N}\left(\mu_{y}, \sigma_{y}^2\right)$, respectively. The probability density functions of these two Gaussian distributions are:
\begin{equation}
P(x)_{x \sim p_c\left(\hat{s}\mid y\right)}=\frac{1}{\sqrt{2 \pi \sigma^2}} \exp \left(-\frac{\left(x- y\right)^2}{2 \sigma^2}\right),
\end{equation}
\begin{equation}
Q(x)_{x \sim p_{\omega}\left(\hat{y}\mid\hat{z}\right)}=\frac{1}{\sqrt{2 \pi \sigma_{y}^2}} \exp \left(-\frac{\left(x- \mu_{y}\right)^2}{2 \sigma_{y}^2}\right).
\end{equation}

The KL divergence between two probability distributions $p_c\left(\hat{s}\mid y\right)$ and $p_{\omega}\left(\hat{y}\mid\hat{z}\right)$ can be represented as:
\begin{equation}
\begin{aligned}
&~~~~~~D_{\text{KL}}\left(p_c\left(\hat{s}\mid y\right) \| p_{\omega}\left(\hat{y}\mid\hat{z}\right)\right)\\
&=\int_{-\infty}^{\infty} P(x) \log \left(\frac{P(x)}{Q(x)}\right) d x \\
&=\int_{-\infty}^{\infty} P(x) \log \left(\frac{\frac{1}{\sqrt{2 \pi \sigma^2}} \exp \left(-\frac{\left(x-y\right)^2}{2 \sigma^2}\right)}{\frac{1}{\sqrt{2 \pi \sigma_y^2}} \exp \left(-\frac{\left(x-\mu_x\right)^2}{2 \sigma_y^2}\right)}\right) dx \\
%&=\!\!\int_{-\infty}^{\infty}\!\!\!\frac{1}{\sqrt{2 \pi \sigma^2}}\! \exp \!\left(\!\!-\!\!\frac{\left(\!x\!-\!y\!\right)^2}{2 \sigma^2}\!\!\right)\! \log \!\! \left(\!\frac{\frac{1}{\sqrt{2 \pi \sigma^2}} \!\exp \!\left(\!-\!\frac{\left(\!x\!-\!y\!\right)^2}{2 \sigma^2}\!\right)}{\frac{1}{\sqrt{2 \pi \sigma_y^2}}\! \exp \!\left(\!-\!\frac{\left(\!x\!-\!\mu_x\!\right)^2}{2 \!\sigma_y^2}\!\right)}\!\!\right) dx \\
&=\int_{-\infty}^{\infty} P(x) \log \left(\left(\frac{\sigma_y}{\sigma}\right)+\frac{\left(x-y\right)^2}{2 \sigma^2}-\frac{\left(x-\mu_y\right)^2}{2 \sigma_y^2}\right)dx \\
&=\log\left(\frac{\sigma_{y}}{\sigma}\right)+\frac{\sigma^2+\left(\mu_{y}-y\right)^2}{2 \sigma^2_{y}}-\frac{1}{2}.
\end{aligned}
\end{equation}
\section{Proof of Theorem 2}
\allowdisplaybreaks[4]
To enhance the stability of the loss function and reduce dependency on the direct outputs of the diffusion model $\boldsymbol{y}^{\prime}$, we introduce an intermediary $\boldsymbol{s}^{\prime}$. This intermediary serves as a bridge between the model's output and the final loss calculation, facilitating more robust training by mitigating the direct impact of fluctuations or noise in the model's predictions. By optimizing the loss function through this intermediary, we effectively decouple the immediate influence of $\boldsymbol{y}^{\prime}$, allowing for more stable gradient updates. The revised loss function can be expressed as follows:
\begin{equation}
\begin{aligned}
& \mathbb{E}\!\left[\!\left(\!{\boldsymbol{y}^{\prime}\!-\!\boldsymbol{y}}\right)^2\!\right] \\
=& \mathbb{E}\!\left[\!\left(\!{\boldsymbol{y}^{\prime}\!-\!\boldsymbol{s}^{\prime}}\!+\!{\boldsymbol{s}^{\prime}\!-\!\boldsymbol{y}}\right)^2\!\right] \\
=& \mathbb{E}\!\left[\!\left(\!{\boldsymbol{y}^{\prime}\!-\!\boldsymbol{s}^{\prime}}\right)^2\!+2\left(\!\boldsymbol{y}^{\prime}\!-\! \boldsymbol{s}^{\prime}\right)\! \otimes \left(\!\boldsymbol{s}^{\prime}\!-\!\boldsymbol{y}\right)\!+\!\left(\!{\boldsymbol{s}^{\prime}\!-\! \boldsymbol{y}}\right)^2\!\right]\!,
\end{aligned}
\label{loss_function}
\end{equation}
where $\boldsymbol{y}^{\prime}\!\sim\!\mathcal{N}\!\left(\!\boldsymbol{\mu}_{\boldsymbol{\phi}}\!\left(\!\hat{\boldsymbol{y}}_{1}\!\right)\!,\!\boldsymbol{\Sigma}_{\boldsymbol{\phi}}\!\right)$, $\boldsymbol{s}^{\prime}\!\sim\!\mathcal{N}\!\left(\!\boldsymbol{\mu}_{\boldsymbol{\phi}}\!\left(\!\hat{\boldsymbol{s}}_{1}\!\right)\!,\!\boldsymbol{\Sigma}_{\boldsymbol{\phi}}\!\right)$. Here, $\hat{\boldsymbol{y}}_{t}$ and $\hat{\boldsymbol{s}}_{t}$ represent the $t$-th step feature of $\hat{\boldsymbol{y}}$ and $\hat{\boldsymbol{s}}$ in the reverse process of diffusion, respectively. Additionally, $\boldsymbol{y}^{\prime}$ and $\boldsymbol{s}^{\prime}$ denote the 0-th step feature $\hat{\boldsymbol{y}}_{0}$ and $\hat{\boldsymbol{s}}_{0}$, respectively. We assume that $\left(\boldsymbol{y}^{\prime} - \boldsymbol{s}^{\prime}\right)$ and $\left(\boldsymbol{s}^{\prime} - \boldsymbol{y}\right)$ are independently and identically distributed (i.i.d.). Thus, Eqn. (\ref{loss_function}) can be represented as:
%Here, we assume that $\left({\boldsymbol{z}_{st}^{\prime}  - \boldsymbol{z}_{te}^{\prime}} \right)$ and $\left({\boldsymbol{z}_{te}^{\prime}  - \boldsymbol{z}}_{te}\right)$ are independently identically distributed (i.i.d.).:
% \frac{1}{\bar{\alpha}_{u}} 
\begin{equation}
\begin{aligned}
&\mathbb{E}\left[\left({\boldsymbol{y}^{\prime}  - \boldsymbol{y}} \right)^2\right] \\
=& \mathbb{E}\!\left[\left({\boldsymbol{y}^{\prime}\!-\!\boldsymbol{s}^{\prime}}\right)^2\right]\!+2\mathbb{E}\left[\left(\!\boldsymbol{y}^{\prime}\!-\! \boldsymbol{s}^{\prime}\right)\right]\mathbb{E}\left[\left(\!\boldsymbol{s}^{\prime}\!-\!\boldsymbol{y}\right)\right]\!+\mathbb{E}\left[\left(\!{\boldsymbol{s}^{\prime}\!-\! \boldsymbol{y}}\right)^2\!\right] \\
=& \mathbb{E}\!\left[\left({\boldsymbol{\mu}_{\boldsymbol{\phi}}\!\left(\!\hat{\boldsymbol{y}}_{1}\!\right)-\!\boldsymbol{\mu}_{\boldsymbol{\phi}}\!\left(\!\hat{\boldsymbol{s}}_{1}\!\right)}\right)^2\right]\!+ C_2  \\
& ~~~~~+ 2\mathbb{E}\left[\left(\!\boldsymbol{y}^{\prime}\!-\! \boldsymbol{s}^{\prime}\right)\right]\mathbb{E}\left[\left(\!\boldsymbol{\mu}_{\boldsymbol{\phi}}\!\left(\!\hat{\boldsymbol{s}}_{1}\!\right)\!-\!\boldsymbol{y}\right)\right]\!+\mathbb{E}\left[\left(\!{\boldsymbol{s}^{\prime}\!-\! \boldsymbol{y}}\right)^2\!\right], \\
\end{aligned}
\end{equation}
where $\boldsymbol{\mu}_{\boldsymbol{\phi}}\!\left(\!\hat{\boldsymbol{y}}_{t}\!\right)\!\sim\!\mathcal{N}\!\left(\!\frac{\hat{\boldsymbol{y}}_{t}}{\sqrt{\alpha_t}}\!, \!\frac{\left(\!1\!-\!\alpha_t\!\right)^2}{\alpha_t\!(\!1\!-\!\bar{\alpha}_t\!)}\! \right)\!$, $\!\boldsymbol{\mu}_{\boldsymbol{\phi}}\!\left(\!\hat{\boldsymbol{s}}_{t}\!\right)\!\sim\!\mathcal{N}\!\left(\!\frac{\hat{\boldsymbol{s}}_{t}}{\sqrt{\alpha_t}}\!,\!\frac{\left(\!1\!-\!\alpha_t\!\right)^2}{\alpha_t\!(\!1\!-\!\bar{\alpha}_t\!)}\!\right)$, according to Eqn. (\ref{reverse}). $\mathbb{E}\left[\left(\boldsymbol{s}^{\prime}  - \boldsymbol{y}\right)^2 \right] = C_1$, since the parameters of the uncompressed generator are frozen. Eqn. (\ref{loss_function}) can be further  expressed as:
\begin{equation}
\begin{aligned}
&\mathbb{E}\left[\left({\boldsymbol{y}^{\prime}  - \boldsymbol{y}} \right)^2\right] \\
=& \mathbb{E}\!\left[\left({\boldsymbol{\mu}_{\boldsymbol{\phi}}\!\left(\!\hat{\boldsymbol{y}}_{1}\!\right)-\!\boldsymbol{\mu}_{\boldsymbol{\phi}}\!\left(\!\hat{\boldsymbol{s}}_{1}\!\right)}\right)^2\right]\!+ C_2  \\
& ~~~~~+ 2\mathbb{E}\left[\left(\!\boldsymbol{y}^{\prime}\!-\! \boldsymbol{s}^{\prime}\right)\right]\mathbb{E}\left[\left(\!\boldsymbol{\mu}_{\boldsymbol{\phi}}\!\left(\!\hat{\boldsymbol{s}}_{1}\!\right)\!-\!\boldsymbol{y}\right)\right]\!+\mathbb{E}\left[\left(\!{\boldsymbol{s}^{\prime}\!-\! \boldsymbol{y}}\right)^2\!\right] \\
=& \frac{1}{\alpha_1}\mathbb{E}\!\left[\left(\hat{\boldsymbol{y}}_{1}-\hat{\boldsymbol{s}}_{1}\!\right)^2\right]\!+ C_3  \\
& ~~~~~+ \frac{2}{\sqrt{\alpha_1}}\mathbb{E}\left[\left(\!\boldsymbol{y}^{\prime}\!-\! \boldsymbol{s}^{\prime}\right)\right]\mathbb{E}\left[\left(\!\hat{\boldsymbol{s}}_{1}\!-\!\boldsymbol{y}\right)\right]\!+\mathbb{E}\left[\left(\!{\boldsymbol{s}^{\prime}\!-\! \boldsymbol{y}}\right)^2\!\right] \\
=& \frac{1}{\bar{\alpha}_{u}}\mathbb{E}\!\left[\left(\hat{\boldsymbol{y}}_{u}-\hat{\boldsymbol{s}}_{u}\!\right)^2\right]\!+ C_4  \\
& ~~~~~+ \frac{2}{\sqrt{\bar{\alpha}_{u}}}\mathbb{E}\left[\left(\!\boldsymbol{y}^{\prime}\!-\! \boldsymbol{s}^{\prime}\right)\right]\mathbb{E}\left[\left(\!\hat{\boldsymbol{s}}_{u}\!-\!\boldsymbol{y}\right)\right]\!+\mathbb{E}\left[\left(\!{\boldsymbol{s}^{\prime}\!-\! \boldsymbol{y}}\right)^2\!\right] \\
= &\underbrace{\mathbb{E}\left[\left(\hat{\boldsymbol{y}} - \hat{\boldsymbol{s}}\right)^2\right] + C_5}_{\mathbb{E}\left[\left({\boldsymbol{y}^{\prime}-\boldsymbol{s}^{\prime}}\right)^2\right]} + \underbrace{2 \mathbb{E}\left[\left(\hat{\boldsymbol{y}} - \hat{\boldsymbol{s}}\right)\right] \underbrace{\mathbb{E}\left[ \left(\hat{\boldsymbol{s}} - \boldsymbol{y}\right)\right]}_{\mathbb{E}\left[\left(\sigma \cdot \boldsymbol{\epsilon}\right)\right]=0}}_{2 \mathbb{E}\left[\left(\boldsymbol{y}^{\prime}- \boldsymbol{s}^{\prime}\right)\otimes \left(\boldsymbol{s}^{\prime}-\boldsymbol{y}\right)\right]} + C_1 \\
= & \mathbb{E}\left[\left(\hat{\boldsymbol{y}} - \hat{\boldsymbol{s}}\right)^2\right] + C.
\end{aligned}
\end{equation}
\bibliographystyle{IEEEtran}
\bibliography{IEEEabrv,reference}

\end{document}